\documentclass[journal,twocolumn]{IEEEtran}

\usepackage{times}
\usepackage{epsfig}
\usepackage{graphicx}
\usepackage{amsmath}
\usepackage{amssymb}
\usepackage{algorithmic}
\usepackage[linesnumbered,ruled,vlined]{algorithm2e}
\usepackage{multirow}
\usepackage{booktabs}
\usepackage{siunitx}
\usepackage{bm}
\usepackage{amsfonts}
\usepackage{caption} 
\usepackage{CJK}
\newif\ifdraft\drafttrue
\ifdraft

\fi
\usepackage[pagebackref=true,breaklinks=true,letterpaper=true,colorlinks,bookmarks=false]{hyperref}
\ifCLASSINFOpdf
\else
\fi
\begin{document}
\begin{CJK*}{UTF8}{gkai}
\title{Discriminative Noise Robust Sparse Orthogonal Label Regression-based Domain Adaptation}

\author{Lingkun~Luo,
	Liming~Chen,~\IEEEmembership{Senior~Member,~IEEE,}
	and~Shiqiang~Hu,
	\thanks{K. Luo, S. Qiang are with School of Aeronautics and Astronautics, Shanghai Jiao Tong University, 800 Dongchuan Road, Shanghai, China e-mail: lolinkun@gmail.com, sqhu@sjtu.edu.cn.}
	\thanks{L. Chen is with LIRIS, CNRS UMR 5205, 	Ecole Centrale de Lyon, 36 avenue Guy de Collongue, Ecully,  France e-mail: (liming.chen)@ec-lyon.fr.}
	\thanks{Manuscript received May 15, 2018.}}

\markboth{Journal of \LaTeX\ 2018}%
{Shell \MakeLowercase{\textit{et al.}}: Bare Demo of IEEEtran.cls for IEEE Journals}

\maketitle

\begin{abstract}

Domain adaptation (\textbf{DA}) aims to enable a learning model trained from a source domain to generalize well on a target domain, despite the mismatch of data distributions between the two domains. State-of-the-art DA methods have so far focused on the search of a latent shared feature space where source and target domain data can be aligned either statistically and/or geometrically.  In this paper, we propose a novel unsupervised DA method, namely \textit{D}iscriminative Noise Robust Sparse \textit{O}rthogonal \textit{L}abe\textit{l} Regression-based \textit{D}omain \textit{A}daptation (\textbf{DOLL-DA}). The proposed DOLL-DA derives from a novel integrated model which searches a shared feature subspace where source and target domain data are, through optimization of some \textit{repulse force} terms,  discriminatively aligned statistically, while at same time regresses orthogonally data labels thereof using a label embedding trick. Furthermore, in minimizing  a novel \textit{Noise Robust Sparse Orthogonal Label Regression}(NRS\_OLR) term, the proposed model explicitly accounts for data outliers to avoid negative transfer and introduces the property of sparsity when regressing data labels. We carry out comprehensive experiments in comparison with 32 state of the art DA methods using 8 standard DA benchmarks and 49 cross-domain image classification tasks. The proposed DA method demonstrates its effectiveness and consistently outperforms the state-of-the-art DA methods with a margin which  reaches 17 points on the CMU PIE dataset. To gain insight into the proposed \textbf{DOLL-DA}, we also derive three additional DA methods based on three partial models from the full model, namely \textbf{OLR}, \textbf{CDDA+}, and \textbf{JOLR-DA}, highlighting the added value of 1) discriminative statistical data alignment; 2)Noise Robust Sparse Orthogonal Label Regression; and 3) their joint optimization through the full DA model. In addition, we also perform time complexity and an in-depth empiric analysis of the proposed DA method in terms of its sensitivity \textit{w.r.t.} hyper-parameters, convergence speed, impact of the base classifier and random label initialization as well as performance stability \textit{w.r.t.} target domain data being used in traing.

\end{abstract}

\begin{IEEEkeywords}
Domain adaptation, Transfer Learning, Visual classification, Noise Robust Sparse Orthogonal Label Regression
\end{IEEEkeywords}

%
\IEEEpeerreviewmaketitle

\section{Introduction}
\label{Introduction}


Traditional machine learning tasks assume that both training and testing data are drawn from a same data distribution \cite{pan2010survey,7078994}. However, in many real-life applications, due to different factors as diverse as sensor difference, lighting changes, viewpoint variations, \textit{etc.}, data from a target domain may have a different data distribution with respect to the labeled data in a source domain where a predictor can not be reliably learned. On the other hand, manually labeling enough target domain data for the purpose of training an effective predictor can be very expensive, tedious and thus prohibitive.  


Domain adaptation (DA) \cite{pan2010survey,7078994, Ying_TL_Survey_10.1145/3379344} aims to leverage possibly abundant labeled data from a \textit{source} domain to learn an effective predictor for data in an unseen domain, namely \textit{target} domain, despite the data distribution discrepancy between the source and target. While DA can be \textit{semi-supervised} by assuming that a certain amount of labeled data is available in the target domain, in this paper we are interested in \textit{unsupervised} DA where we assume that no labels are available for target domain data.

While there exists an increasing number of deep learning-based unsupervised DA methods, we focus in this paper on \textit{shallow} DA methods as they are easier to train and can provide insights into the design decisions of deep DA methods.  The relationships between \textit{shallow} and \textit{deep} DA methods will be discussed in depth in Sect.\ref{Related Work} on related works. State of the art \textit{shallow}  DA methods can be categorized into \textit{instance}-based \cite{pan2010survey,donahue2013semi}, \textit{feature}-based \cite{Busto_2017_ICCV,long2013transfer,DBLP:journals/tip/XuFWLZ16}, or \textit{classifier}-based. Classifier-based DA is widely applied in semi-supervised DA as it aims to fit a classifier trained on  source domain data to target domain data through adaptation of its parameters, and thereby require some labels in the target domain\cite{tang2017visual}.  The instance-based approach generally assumes that 1) the conditional distributions of source and target domain are identical\cite{Zhang_2017_CVPR}, and 2) certain portion of the data in the source domain can be reused\cite{pan2010survey} for learning in the target domain through re-weighting. Feature-based adaptation\cite{sun2016return,DBLP:conf/cvpr/HerathHP17,ganin2016domain,tzeng2017adversarial, luo2020discriminative} relaxes such a strict assumption and only requires that there exists a mapping from the input data space to a latent shared feature representation space. This latent shared feature space captures the information necessary for training classifiers for source and target tasks. In this paper, we propose a novel \textit{hybrid} DA method using both feature adaptation and classifier optimization.


A common method to approach feature adaptation is to seek a shared latent subspace between the source and target domain \cite{7078994,Busto_2017_ICCV} via optimization.  State-of-the-art  features three main lines of approaches, namely, data geometric structure alignment-based (\textbf{DGSA}), data distribution centered (\textbf{DDC}) or their hybridization.  \textbf{DGSA} based approaches \cite{sun2016return,DBLP:journals/ijcv/ShaoKF14,DBLP:journals/tip/XuFWLZ16}  seek a subspace where source and target data can be well aligned and interlaced in preserving inherent hidden geometric data structure via low rank constraint and/or sparse representation.  \textbf{DDC} methods \cite{DBLP:journals/corr/abs-1712-10042, DBLP:conf/icml/LongZ0J17, liang2018aggregating,lu2018embarrassingly} aim to search a latent subspace where the discrepancy between the source and target data distributions is minimized, via various distances, \textit{e.g.}, Bregman divergence\cite{si2010bregman}, Geodesic distance\cite{gong2012geodesic}, Wasserstein distance\cite{courty2017joint, courty2017optimal} or Maximum Mean Discrepancy\cite{gretton2007kernel} (MMD). The most popular distance is MMD due to its simplicity and solid theoretical foundation. 

Our previously proposed DA method, namely \textbf{DGA-DA}\cite{luo2020discriminative}, is a hybridization of \textbf{DDC} and \textbf{DGSA} approaches. \textbf{DGA-DA} leverages the advantages of both \textbf{DDC} and \textbf{DGSA} methods and demonstrates a state of the art performance on a number of DA benchmarks. \textbf{DGA-DA} relies upon the analysis of a cornerstone theoretical result in DA \cite{ben2007analysis,ben2010theory,kifer2004detecting}, which estimates an error bound of a learned hypothesis $h$ on a  target domain as follows:  
\begin{equation}\label{eq:bound}
	\resizebox{0.90\hsize}{!}{%
		$\begin{array}{l}
		{e_{\cal T}}(h) \le {e_{\cal S}}(h) + {d_{\cal H}}({{\cal D}_{\cal S}},{{\cal D}_{\cal T}})+ \\
		\;\;\;\;\;\;\;\; \min \left\{ {{{\cal E}_{{{\cal D}_{\cal S}}}}\left[ {\left| {{f_{\cal S}}({\bf{x}}) - {f_{\cal T}}({\bf{x}})} \right|} \right],{{\cal E}_{{{\cal D}_{\cal T}}}}\left[ {\left| {{f_{\cal S}}({\bf{x}}) - {f_{\cal T}}({\bf{x}})} \right|} \right]} \right\}
		\end{array}$}
\end{equation}

Our previously proposed \textbf{DGA-DA} provides a unified framework which jointly optimizes term 2 in Eq.(\ref{eq:bound}) as \textbf{DDC-DA} methods when aligning data distributions, and term 3 in Eq.(\ref{eq:bound}) as \textbf{DGSA-DA} approaches when performing label inference through the underlying data geometric structure. It further introduces a \textit{repulsive force}(\textbf{RF}) term using both source and target domain data when seeking the latent feature space and makes the proposed DA method discriminative.  In this paper, we go one step further and propose a novel DA method, namely \textbf{D}iscriminative Noise Robust Sparse \textbf{O}rthogonal \textbf{L}abe\textbf{l} Regression-based \textbf{D}omain \textbf{A}daptation (\textbf{DOLL-DA}), which optimizes at the same time the three terms of the right-hand of Eq.(\ref{eq:bound}), including in particular the first term on classification error in Eq.(\ref{eq:bound}), when seeking a discriminative latent feature space. 

\begin{figure*}[h!]
	\centering
	\includegraphics[width=1\linewidth]{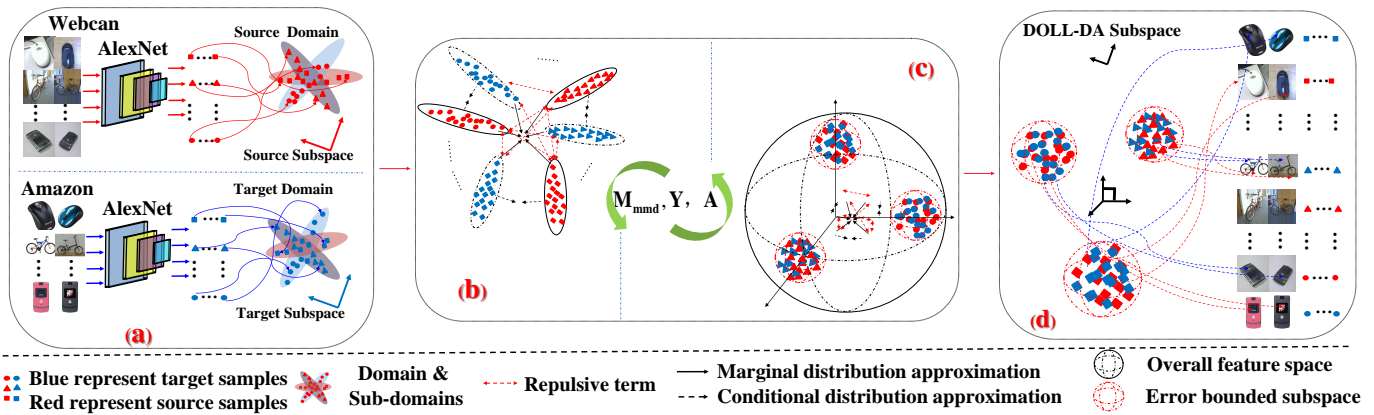}

	\caption {Illustration of the proposed \textbf{DOLL-DA} method. Fig.\ref{fig:diff} (a): source domain data and target domain data, \textit{e.g.},   mouse, bike, smartphone images, with different distributions and inherent hidden data geometric structures between the source in red  and the target in blue. Samples of different class labels are represented by different geometrical shapes, \textit{e.g.}, round, triangle and square; 	Fig.\ref{fig:diff} (b) illustrates \textbf{DOLL-DA} which aligns data distributions closely yet discriminatively through the use of the nonparametric distance, \textit{i.e.}, Maximum Mean Discrepancy (MMD). Fig.\ref{fig:diff} (c): accounts for well regularization of the designed \textbf{MMD} distances, which intends to enable the different sub-domains, \textit{w.r.t}, its orthogonal subspace meanwhile cares about the noise data; In \textbf{DOLL-DA}, the optimized common subspace ${\bf{A}}$ and label matrix ${\bf{Y}}$ are updated iteratively within the processes in Fig.\ref{fig:diff} (b-c);  Fig.\ref{fig:diff} (d): the achieved latent joint subspace where both marginal and class conditional data distributions are aligned discriminatively through  the well proposed orthogonal regularization; Furthermore, noise data are well accounted as well as the hidden manifold structure, the formal one intends to avoide the negative transfer while the latter one improve the transferability of the learned model based on the source domain.} 
	\label{fig:diff}
\end{figure*}

Specifically, the proposed \textbf{DOLL-DA} derives from an integrated DA model which: \textbf{1)} searches a shared feature subspace where the source and target data distributions are discriminatively aligned using an improved \textit{repulsive force} (\textbf{RF}) term added to the  \textbf{MMD} constraints, thereby optimizing the second term in Eq.(\ref{eq:bound}) and indirectly improving its first term; \textbf{2)} makes use an embedding trick to immerse data labels into the shared feature space and projects each data sample within the vicinity of its label vector orthogonal to other ones, thereby further regularizing the improved \textbf{MMD} constraints and avoiding potential contradictions among sub-domains when \textit{repulsive force} is applied in the search of the shared feature subspace; \textbf{3)} linearly regresses data labels in the shared feature subspace, thereby further explicitly optimizes the first term of the right-hand in Eq.(\ref{eq:bound}). Moreover, data outliers are accounted for and the property of sparsity in label regression is introduced to circumvent negative transfer and over-fitting, leading to a novel \textit{Noise Robust Sparse Orthogonal Label Regression} (\textbf{NRS\_OLR}) term in our DA model;  \textbf{4)} leverages the true labels available in the source domain and ensures a \textit{label consistency} between the source and target domain through an iterative integrated linear label regression, thereby minimizing the third term of Eq.(\ref{eq:bound}); Fig.\ref{fig:diff} depicts the general framework of the proposed \textbf{DOLL-DA} method.

To sum up, the contributions of this paper are as follows: 

\begin{itemize}
	\item Improved \textit{repulsive force} (\textbf{RF})-based MMD constraints are introduced to enable  discriminative alignment of data distributions between the source and the target domain. 

	\item Orthogonal label subspace is proposed through a label embedding trick to further regularize the improved  \textbf{RF}-based  \textbf{MMD} constraints using a novel \textit{Orthogonal Label Regression} (\textbf{ORL}) constraint, thereby circumventing potential conflicts which could arise when optimizing the improved \textbf{RF}-based MMD constraints and further enhancing the discriminative power of the proposed DA method.

	\item The hypothesis for classifying the target domain data is learned simultaneously through a single feature projection matrix ${\bf{A}}$ when aligning discriminatively the source and target domain data in the shared feature subspace.  Furthermore, a property of sparsity is introduced through a ${l_{2,1}}$-norm constraint on ${\bf{A}}$ when regressing data labels and data outliers are also accounted for within the model, leading to a \textit{Noise Robust Sparse Orthogonal Label Regression}(\textbf{NRS\_ORL}) term to ensure the proposed DA model to avoid negative transfer as well as overfitting.  
	
	\item A novel generalized power iteration method is introduced to solve the optimization problem of the full integrated DA model, resulting in the proposed \textbf{DOLL-DA}. Furthermore, we perform time complexity analysis and also derive three additional DA methods based on three partial models from the full integrated DA in order to highlight the individual contribution of \textbf{RF} and \textbf{NRS\_OLR} term, respectively, as well as their added value when they are jointly optimized.

	
	

	\item We perform extensive experiments on 49 image classification DA tasks using 8 popular \textbf{DA} benchmarks and demonstrate the effectiveness of the proposed \textbf{DOLL-DA} which consistently outperforms thirty state-of-the-art \textbf{DA} algorithms with a margin which can reach 17 points. Moreover, we also carry out in-depth analysis of the proposed \textbf{DA} method, in particular \textit{w.r.t.} their hyper-parameters, convergence speed, the choice of the base classifier, random label initialization and impact of the quantity of target domain data for performance stability.

\end{itemize}

The article  is organized as follows. Sect.\ref{Related Work} discusses the related work. Sect.\ref{The proposed method} presents the method. Sect.\ref{Experiments} benchmarks the proposed DA method and provides in-depth analysis. Sect.\ref{Conclusion} draws the conclusion.

\section{Related Work}
\label{Related Work}

Last years have seen \textbf{DA} techniques applied to  multiple computer vision applications. State-of-the-art has so far featured two main research streams: 1) Shallow \textbf{DA}; 2) Deep \textbf{DA}. They are overviewed in Sect.\ref{subsect: Shallow DA} and sect.\ref{subsect: Deep DA}, respectively, and discussed in comparison with the proposed \textbf{DOLL-DA} in sect.\ref{subsection:Discussion}.

\subsection{Shallow Domain Adaptation}
\label{subsect: Shallow DA}

\subsubsection{Feature-based DA}
\label{feature-based DA}
The rationale of the \textit{feature}-based domain adaptation is to assume a shared latent feature space between the source and target domain which is searched  in narrowing the existing distribution discrepancies across the domains. A popular strategy for searching such a shared latent feature space is to embrace the dimensionality reduction and propose to explicitly minimize some predefined distance measure to reduce the mismatch between source and target in terms of marginal distribution \cite{4967588} \cite{pan2011domain}, or conditional distribution \cite{long2013transfer}. These methods can be further distinguished based on whether they incorporate some form of data discriminativeness or not in the search of such a shared latent feature space.




\textbf{Nondiscriminative distribution alignment (NDA):}
\textbf{NDA} strategies propose to align the marginal and conditional distributions across the source and target domains in reducing different data distribution distance measurements,\textit{ e.g.}, Bregman Divergence \cite{4967588},  Wasserstein distance \cite{courty2017joint, courty2017optimal}, \textit{Maximum Mean Discrepancy} (MMD)  to explicitly shrink the cross-domain divergence of marginal data distributions  \cite{pan2011domain}, and both the  marginal and conditional data distributions \cite{long2013transfer}. The obvious drawback of \textbf{NDA} is that it ignores the discriminative knowledge among different labeled source sub-domains , thereby increasing the burden of the required classifier.  



\textbf{Discriminative distribution alignment (DDA):}
\textbf{DDA} methods improve \textbf{NDA} ones by explicitly leveraging the discriminative information in source domain data labeled into different sub-domains. \textbf{ILS}\cite{herath2017learning} learns a discriminative  latent space using Mahalanobis metric and makes use of Riemannian optimization strategy to match statistical properties across different domains. \cite{lu2018embarrassingly} adapts linear discriminant analysis (\textbf{LDA}) and leverages the discriminative information from the target domain to estimate the common feature space.  \textbf{OBTL}\cite{karbalayghareh2018optimal} proposes bayesian transfer learning based domain adaptation, which explicitly discusses the relatedness across different sub-domains.  \textbf{SCA}\cite{DBLP:journals/pami/GhifaryBKZ17} achieves discriminativeness in optimizing the interplay of the between and within-class scatters. Our proposed \textbf{DGA-DA}\cite{luo2020discriminative} also  introduces a specific \textit{repulsive force} term to capture the data discriminativeness. However, our \textbf{DGA-DA} further cares about the underlying data manifold structure when performing label inference. However, despite improvement over \textbf{NDA} methods, \textbf{DDA} methods  rely upon temporary and unreliable pseudo labels of target domain data to capture the repulsive force in the target domain, and thereby can mislead the search of an optimized shared discriminative latent space. 


\subsubsection{Subspace alignment-based DA}
In line with \cite{DBLP:conf/iccv/FernandoHST13}, an increasing number of DA methods, \textit{e.g.}, \cite{DBLP:journals/corr/LuoWHC17,DBLP:journals/ijcv/ShaoKF14,DBLP:journals/tip/XuFWLZ16,sun2016return,DBLP:journals/tnn/DingF18},  emphasize the importance of aligning the underlying data subspace and manifold structures between the source and the target domain for effective DA. In these methods, low-rank and sparse constraints are introduced into DA to extract a low-dimension feature subspace where target samples can be sparsely reconstructed from source samples \cite{DBLP:journals/ijcv/ShaoKF14}, or interleaved by source samples \cite{DBLP:journals/tip/XuFWLZ16},  thereby aligning the geometric structures of the underlying data manifolds. A few recent DA methods, \textit{e.g.}, \textbf{RSA-CDDA}\cite{DBLP:journals/corr/LuoWHC17}, \textbf{JGSA}\cite{Zhang_2017_CVPR}, further propose unified frameworks to reduce the shift between domains both statistically and geometrically. \textbf{HCA}\cite{liu2019homologous} improves \textbf{JGSA} using a  homologous constraint on the two transformations for the source and target domains,  respectively, to make the transformed domains related and hence alleviate negative domain adaptation.


However, in light of the upper error bound as defined in Eq.(\ref{eq:bound}), we can see that subspace alignment based DA methods account for the underlying data geometric structure and expect but without theoretical guarantee the alignment of discriminative data distributions. Our proposed \textbf{DOLL-DA} improves subspace alignment-based DA by jointly aligning the data distributions discriminatively  and enhancing the hidden manifold structure through label regression to regress different sub-domains \textit{w.r.t} its orthogonal label subspace. 




\subsection{Deep Domain Adaptation}
\label{subsect: Deep DA}
Recently, \textbf{DA} has been intensively investigated under the paradigm of deep learning (\textbf{DL}), and has featured the following two main approaches.

\subsubsection{Statistic matching-based DA}
 These methods aim to reduce the divergence across  domains using statistic measurements incorporated into \textbf{DL} frameworks. \textbf{DAN}\cite{long2015learning} reduces the marginal distribution divergence in incorporating the multi-kernel MMD loss on the fully connected layers of AlexNet. \textbf{JAN}\cite{DBLP:conf/icml/LongZ0J17} improves \textbf{DAN} by jointly decreasing the divergence of both the marginal and conditional distributions.  \textbf{D-CORAL}\cite{sun2016deep} further introduces the second-order statistics into  the AlexNet\cite{krizhevsky2012imagenet} framework for more effective \textbf{DA} strategy.



\subsubsection{Adversarial loss-based DA}
\label{Adversarial loss-based DA}
These methods make use of \textbf{GAN}\cite{goodfellow2014generative} and propose to align data distributions across domains in making sample features indistinguishable \textit{w.r.t} the domain labels through an adversarial loss on a domain classifier \cite{ganin2016domain,tzeng2017adversarial,pei2018multi}. \textbf{DANN}\cite{ganin2016domain} and \textbf{ADDA}\cite{tzeng2017adversarial} learn a  domain-invariant feature subspace in reducing the marginal distribution divergence.
\textbf{MADA} \cite{pei2018multi} additionally make use of multiple domain discriminators, thereby aligning conditional data distributions. Different from the previous approaches, \textbf{DSN}\cite{bousmalis2016domain} achieves domain-invariant representations in explicitly separating the similarities and dissimilarities in the source and target domains.
\textbf{MADAN}\cite{zhao2019multi} explores knowledge from different multi-source domains to fulfill \textbf{DA} tasks. \textbf{CyCADA}\cite{pmlr-v80-hoffman18a} addresses the distribution divergence using a bi-directional \textbf{GAN} based training framework. 


The main advantage of these \textbf{DL} based \textbf{DA} methods is that they jointly shrink the divergence of data distributions across domains and achieve a discriminative feature representation of data through a single unified end-to-end learning framework. However, they also present the drawback that the discriminative force is merely extracted from the labelled source domain while it could rely upon simultaneously the source and target domains in leveraging the underlying data geometric structures. 
Furthermore, these \textbf{DL} based \textbf{DA} approaches mostly work as a black-box and suffer from interpretability, thereby falling short 
to provide deep insights for further improving \textbf{DA} methods.






\subsection{Discussion}
\label{subsection:Discussion}

Fig.\ref{fig:divergence} compares our proposed \textbf{DOLL-DA} with our previously proposed  DA method, \textbf{DGA-DA}\cite{luo2020discriminative} as well as \textbf{MEDA}\cite{wang2018visual}, and highlights their similarities and differences according to the following 6 properties:


\begin{figure}[h!]
	\centering
	\includegraphics[width=1\linewidth]{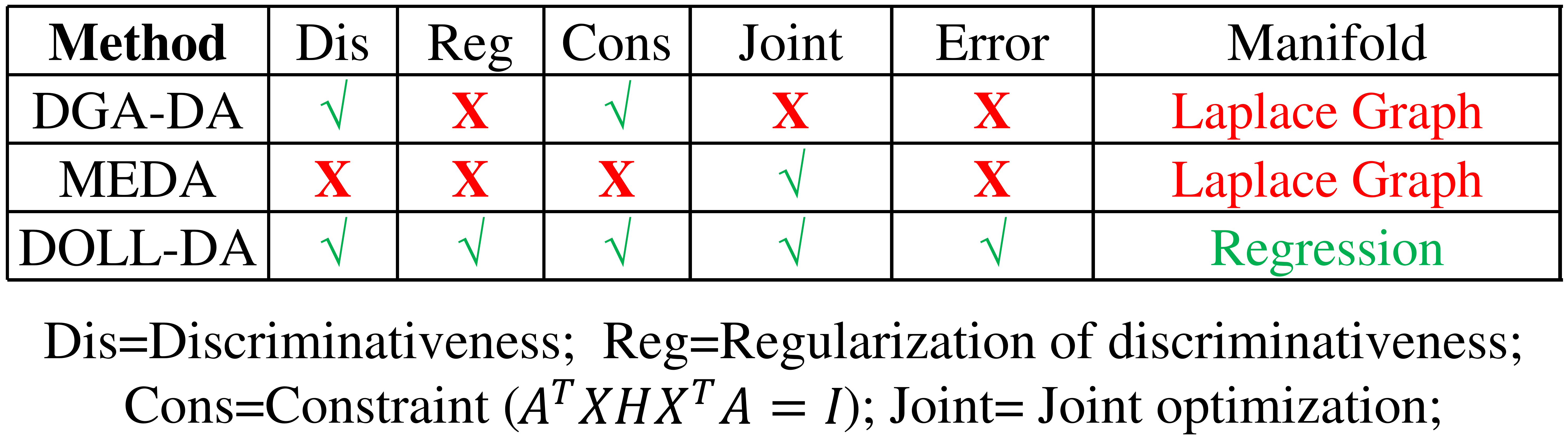}
	\caption{Model comparison} 
	\label{fig:divergence}
\end{figure}

\begin{itemize}
	\item \textbf{Dis}(Discriminativeness): both \textbf{DGA-DA} and \textbf{DOLL-DA} introduce a \textit(\textbf{RF}) term to achieve discriminativeness across domains, while \textbf{MEDA} ignores this merit.  \textbf{DGA-DA} takes into account \textbf{RF} term across domain, while the proposed \textbf{DOLL-DA} improves it in extending it with a \textbf{RF} term within the source domain.   
	
	\item \textbf{Reg}(Regularization): Different from \textbf{DGA-DA} and \textbf{MEDA}, \textbf{DOLL-DA} proposes orthogonal label regression which further regularizes the effectiveness of the improved \textbf{RF}-based \textbf{MMD} constraints, thereby circumventing  potential contradictions when reinforcing these  \textbf{MMD} constraints.
	
	\item \textbf{Cons}(Constraint): Both \textbf{DGA-DA} and \textbf{DOLL-DA} are optimized using the constraint ${{{\bf{A}}^T}{\bf{XH}}{{\bf{X}}^T}{\bf{A}} = {\bf{I}}}$, which removes an arbitrary scaling factor in the embedding and prevents the optimization from collapsing into a subspace of dimension less than the required dimensions. \textbf{MEDA} does not have such a constraint to obtain an analytical solution.

	\item \textbf{Joint}(Joint optimization): Both \textbf{MEDA} and \textbf{DOLL-DA} achieve data distribution alignment and classifier optimization through a unified joint optimization model, whereas the optimization in \textbf{DGA-DA} for data distribution alignment and classifier optimization is carried out separately.

	
	\item \textbf{Error}: \textbf{DOLL-DA} designs an error tolerated subspace to care about the noise in data, therefore decreasing the risk of potential negative transfer.
	
	\item \textbf{Manifold}: both \textbf{MEDA} and \textbf{DGA-DA} capture the underlying data manifold structures for label inference. While effective, they also suffer from the computational burden due to the singular value decomposition. \textbf{DOLL-DA} also cares about data manifold structures, but through computation effective iterative integrated linear label regression.

\end{itemize}

\section{The proposed method}
\label{The proposed method}
Sect.\ref{Notations and Problem Statement} defines the notations and states the DA problem. Sect.\ref{subsection:Formulation} formulates our DA model while Sect.\ref{Solving the model} presents the generalized power iteration method to solve the proposed DA model and derives the algorithm of \textbf{DOLL-DA}. Sect.\ref{Kernelization Analysis} extends \textbf{DOLL-DA} to non-linear problem through kernel mapping.  Sect.\ref{subsection:Time Complexity Analysis} performs time complexity analysis of the proposed \textbf{DOLL-DA}. 

\subsection{Notations and Problem Statement}
\label{Notations and Problem Statement}

Matrices are written as boldface uppercase letters. Vectors are written as boldface lowercase letters. For matrix ${\bf{M}} = ({m_{ij}})$, its $i$-th row is denoted as ${{\bf{m}}^i}$, and its $j$-th column is denoted by ${{\bf{m}}_j}$.  We define the Frobenius norm ${\left\| . \right\|_F}$ and ${l_{2,1}}$ norm as: ${\left\| {\bf{M}} \right\|_F} = \sqrt {\sum {_{i = 1}^n} \sum {_{j = 1}^l} m_{ij}^2} $ and ${\left\| {\bf{M}} \right\|_{2,1}} = \sum {_{i = 1}^n} \sqrt {\sum {_{j = 1}^l} m_{ij}^2} $. A domain $D$ is defined as an l-dimensional feature space $\chi$ and a marginal probability distribution $P(x)$, \textit{i.e.}, $\mathcal{D}=\{\chi,P(x)\}$ with $x\in \chi$.  Given a specific domain $D$, a  task $T$ is composed of a C-cardinality label set $\mathcal{Y}$  and a classifier $f(x)$,\textit{ i.e.}, $T = \{\mathcal{Y},f(x)\}$, where $f({x}) = \mathcal{Q}( y |x)$ can be interpreted as the class conditional probability distribution for each input sample $x$.

In unsupervised domain adaptation, we are given a source domain $\mathcal{D_S}=\{x_{i}^{s},y_{i}^{s}\}_{i=1}^{n_s}$ with $n_s$ labeled samples ${{\bf{X}}_{\cal S}} = [x_1^s...x_{{n_s}}^s]$, which are associated with their class labels ${{\bf{Y}}_S} = {\{ {y_1},...,{y_{{n_s}}}\} ^T} \in {{\bf{\mathbb{R}}}^{{n_s} \times C}}$, and an unlabeled target domain $\mathcal{D_T}=\{x_{j}^{t}\}_{j=1}^{n_t}$ with $n_t$  unlabeled samples ${{\bf{X}}_{\cal T}} = [x_1^t...x_{{n_t}}^t]$, whose labels  ${{\bf{Y}}_T} = {\{ {y_{{n_s} + 1}},...,{y_{{n_s} + {n_t}}}\} ^T} \in {{\bf{\mathbb{R}}}^{{n_t} \times C}}$ are unknown. Here, ${y_i} \in {{\bf{\mathbb{R}}}^c}(1 \le i \le {n_s} + {n_t})$ is a one-vs-all label hot vector in which $y_i^j = 1$ if ${x_i}$ belongs to the $j$-th class, and $0$ otherwise. We  define the data matrix ${\bf{X}} = [{{\bf{X}}_S},{{\bf{X}}_T}] \in {R^{l*n}}$ ($l = feature\;\dim ension$; $n = {n_s} + {n_t}$ ) in packing both the source and target data. The source domain $\mathcal{D_S}$ and target domain $\mathcal{D_T}$ are assumed to be different, \textit{i.e.},  $\mathcal{\chi}_S=\mathcal{{\chi}_T}$, $\mathcal{Y_S}=\mathcal{Y_T}$, $\mathcal{P}(\mathcal{\chi_S}) \neq \mathcal{P}(\mathcal{\chi_T})$, $\mathcal{Q}(\mathcal{Y_S}|\mathcal{\chi_{S}}) \neq \mathcal{Q}(\mathcal{Y_T}|\mathcal{\chi_{T}})$. We also define the notion of \textit{sub-domain}, \textit{i.e.}, class,  denoted as ${\cal D}_{\cal S}^{(c)}$, representing the set of samples in ${{\cal D}_{\cal S}}$ with the class label $c$. It is worth noting that, the definition of sub-domains in the target domain, namely ${\cal D}_{\cal T}^{(c)}$,  requires a base classifier,\textit{ e.g.}, Nearest Neighbor (NN),  to attribute  pseudo labels for samples in ${{\cal D}_{\cal T}}$.


The  maximum mean discrepancy (MMD)  is an effective non-parametric distance-measure  that compares the distributions of two sets of data by mapping the data into Reproducing Kernel Hilbert Space\cite{borgwardt2006integrating} (RKHS). Given two distributions $\mathcal{P}$ and $\mathcal{Q}$, the MMD between $\mathcal{P}$ and $\mathcal{Q}$ is defined as:
\begin{equation}
	\label{eq:MMD}
	Dist(P,Q) = \parallel \frac{1}{n_1} \sum^{n_1}_{i=1} \phi(p_i) - \frac{1}{n_2} \sum^{n_2}_{i=1} \phi(q_i) \parallel_{\mathcal{H}}
\end{equation}
where $P=\{ p_1, \ldots, p_{n_1} \}$ and $Q = \{ q_1, \ldots, q_{n_2} \}$ are two random variable sets from distributions $\mathcal{P}$ and $\mathcal{Q}$, respectively, and $\mathcal{H}$ is a universal RKHS with the reproducing kernel mapping $\phi$: $f(x) = \langle \phi(x), f \rangle$, $\phi: \mathcal{X} \to \mathcal{H}$.

The aim of the proposed DOLL-DA is to search jointly a transformation matrix ${\bf{A}} \in {R^{l*k}}$ projecting discriminatively both the source and target domain data of dimension $l$ into a  latent shared orthogonal feature subspace of dimension $k$ as well as a label regressor while minimizing simultaneously the three terms of the upper error bound in Eq.(\ref{eq:bound}). 

\subsection{Formulation}
\label{subsection:Formulation}

\begin{figure*}[h!]
	\centering
	\includegraphics[width=0.8\linewidth]{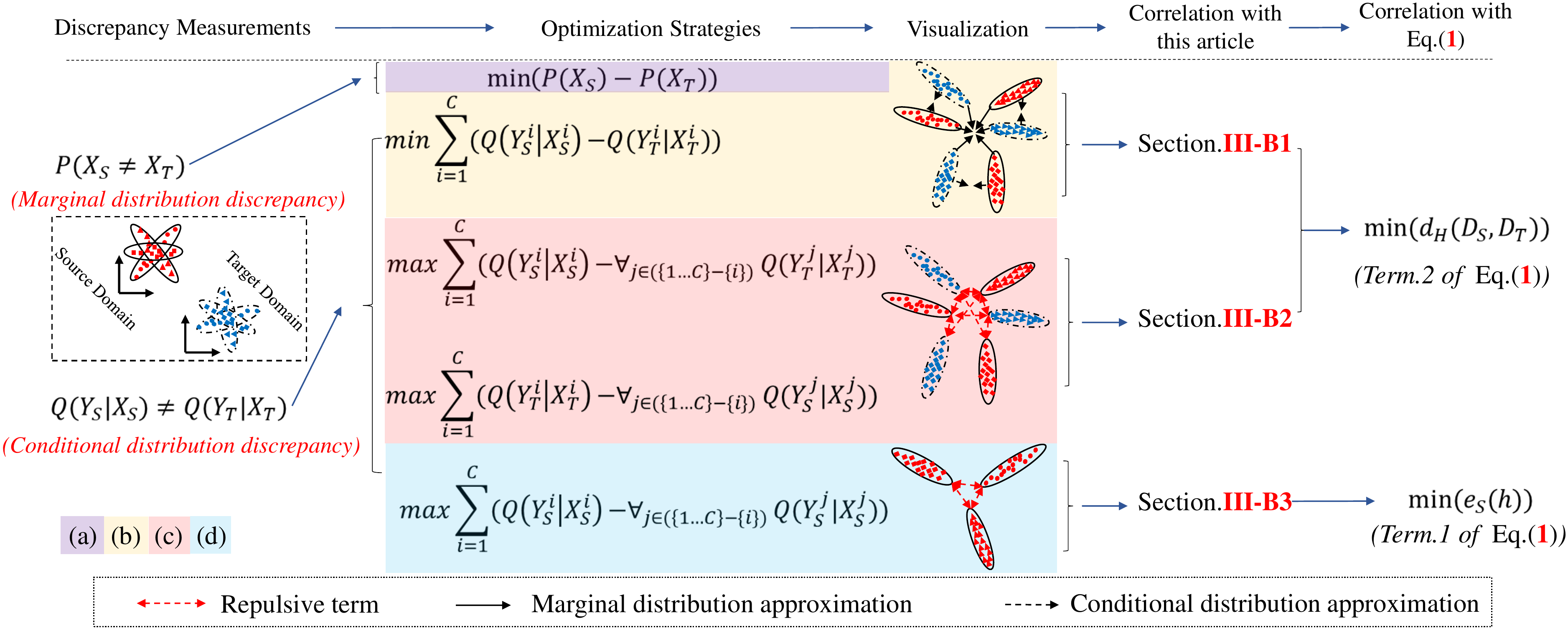}

	\caption {Fig.\ref{fig:div} (a): marginal distribution matching; Fig.\ref{fig:div} (b): conditional distribution matching; Fig.\ref{fig:div} (c): \textit{Repulsive force} term proposed on the source domain; Fig.\ref{fig:div} (d): \textit{Repulsive force} term proposed across the source and target domains. (purple, yellow, blue, and red parts represent Fig.\ref{fig:div} (a), Fig.\ref{fig:div} (b), Fig.\ref{fig:div} (c), and Fig.\ref{fig:div} (d) respectively.)} 
	\label{fig:div}
\end{figure*}

Our final model \textbf{DOLL-DA} (sect.\ref{The final model}) starts from \textbf{JDA} (sect.\ref{subsection:Matching Marginal and Conditional Distributions}), 
which is improved in Sect.\ref{Repulsing interclass data for discriminative DA} and Sect.\ref{Repulsive force term across different domains} for discriminative data distribution alignment (\textbf{DDA})  by leveraging the discriminative knowledge from the source and target domains. Fig.\ref{fig:div} summarizes these steps which aim to minimize ${d_{\cal H}}({{\cal D}_{\cal S}},{{\cal D}_{\cal T}})$ (term.2 of Eq.(\ref{eq:bound})). Sect.\ref{Label Consistent Regression} further introduces an orthogonal label regressor using an embedding trick and accounts for noisy data as well as sparsity in label regression to derive Noise Robust Sparse Orthogonal Label Regression (\textbf{NRS\_OLR}). Sect.\ref{The final model} integrates \textbf{DDA} and \textbf{NRS\_OLR} to achieve our final model and thereby optimizes at the same time the three terms of the right-hand of Eq.(\ref{eq:bound}).  

\subsubsection{Matching Marginal and Conditional Distributions}
\label{subsection:Matching Marginal and Conditional Distributions}

As shown in Fig.\ref{fig:div}.a and Fig.\ref{fig:div}.b, our model starts from \textbf{JDA}, which makes use of MMD in RKHS to measure the distances between the expectations of the source domain/sub-domain and target domain/sub-domain. Specifically, \textbf{1)} The empirical distance of the source and target domains is defined as $Dis{t^{m}}$; \textbf{2)} The conditional distance $Dis{t^{c}}$ is defined as the sum of the empirical distances between sub-domains in ${{\cal D}_{\cal S}}$ and ${{\cal D}_{\cal T}}$ with a same label; \textbf{3)} $Dis{t_{Clo}}$ is defined as the sum of $Dis{t^{m}}$ and $Dis{t^{c}}$.
\begin{equation}\label{eq:JDA}
	\resizebox{0.75\hsize}{!}{%
		$\begin{array}{l}
		Dis{t_{Clo}} = Dis{t^m}({D_S},{D_T}) + Dis{t^c}\sum\limits_{c = 1}^C {({D_S}^c,{D_T}^c)} \\
		\;\;\;\;\;\;\;\;\; = {\left\| {\frac{1}{{{n_s}}}\sum\limits_{i = 1}^{{n_s}} {{{\bf{A}}^T}{x_i} - } \frac{1}{{{n_t}}}\sum\limits_{j = {n_s} + 1}^{{n_s} + {n_t}} {{{\bf{A}}^T}{x_j}} } \right\|^2}\\
		\;\;\;\;\;\;\;\;\; + {\left\| {\frac{1}{{n_s^{(c)}}}\sum\limits_{{x_i} \in {D_S}^{(c)}} {{{\bf{A}}^T}{x_i}}  - \frac{1}{{n_t^{(c)}}}\sum\limits_{{x_j} \in {D_T}^{(c)}} {{{\bf{A}}^T}{x_j}} } \right\|^2}\\
		\;\;\;\;\;\;\;\;\; = tr({{\bf{A}}^T}{\bf{X}}({{\bf{M}}_{\bf{0}}} + \sum\limits_{c = 1}^{c = C} {{{\bf{M}}_c}} ){{\bf{X}}^{\bf{T}}}{\bf{A}})
		\end{array}$}
\end{equation}

\begin{itemize}
	\item {$Dis{t^{m}}({{\cal D}_{\cal S}},{{\cal D}_{\cal T}})$}: where ${{\bf{M}}_0}$ is the MMD matrix  between ${{\cal D}_{\cal S}}$ and ${{\cal D}_{\cal T}}$ with ${{{({{\bf{M}}_0})}_{ij}} = \frac{1}{{{n_s}{n_s}}}}$ if $({{\bf{x}_i},{\bf{x}_j} \in {D_S}})$, ${{{({{\bf{M}}_0})}_{ij}} = \frac{1}{{{n_t}{n_t}}}}$ if $({{\bf{x}_i},{\bf{x}_j} \in {D_T}})$ and ${{{({{\bf{M}}_0})}_{ij}} = 0}$ otherwise.  Thus, the difference between the marginal distributions $\mathcal{P}(\mathcal{X_S})$ and $\mathcal{P}(\mathcal{X_T})$ is reduced when minimizing {$Dis{t^{m}}({{\cal D}_{\cal S}},{{\cal D}_{\cal T}})$}.
	
   \item {$Dis{t^{c}}({{\cal D}_{\cal S}},{{\cal D}_{\cal T}})$}:	where $C$ is the number of classes, $\mathcal{D_S}^{(c)} = \{ {\bf{x}_i}:{\bf{x}_i} \in \mathcal{D_S} \wedge y({\bf{x}_i}) = c\} $ represents the ${c^{th}}$ sub-domain in the source domain, in which $n_s^{(c)} = {\left\| {\mathcal{D_S}^{(c)}} \right\|_0}$ is the number of samples in the ${c^{th}}$ {source} sub-domain. $\mathcal{D_T}^{(c)}$ and $n_t^{(c)}$ are defined similarly for the target domain but using pseudo-labels.  Finally, $\bf M_c$ denotes as the MMD matrix between the sub-domains with labels $c$ in ${{\cal D}_{\cal S}}$ and ${{\cal D}_{\cal T}}$ with ${{{({{\bf{M}}_c})}_{ij}} = \frac{1}{{n_s^{(c)}n_s^{(c)}}}}$ if $({{\bf{x}_i},{\bf{x}_j} \in {D_S}^{(c)}})$, ${{{({{\bf{M}}_c})}_{ij}} = \frac{1}{{n_t^{(c)}n_t^{(c)}}}}$ if $({{\bf{x}_i},{\bf{x}_j} \in {D_T}^{(c)}})$, ${{{({{\bf{M}}_c})}_{ij}} = \frac{{ - 1}}{{n_s^{(c)}n_t^{(c)}}}}$ if $({{\bf{x}_i} \in {D_S}^{(c)},{\bf{x}_j} \in {D_T}^{(c)}\;or\;\;{\bf{x}_i} \in {D_T}^{(c)},{\bf{x}_j} \in {D_S}^{(c)}})$ and ${{{({{\bf{M}}_c})}_{ij}} = 0}$ otherwise. As a consequence, the mismatch of conditional distributions between ${{D_{\cal S}}^c}$ and ${{D_{\cal T}}^c}$ is reduced in minimizing ${Dis{t^{c}}}$.
	
\end{itemize}

\subsubsection{Across domain \textit{Repulsive force} term}
\label{Repulsing interclass data for discriminative DA}

As shown in Fig.\ref{fig:div}.(a,b), sect.\ref{subsection:Matching Marginal and Conditional Distributions} merely cares about shrinking the \textbf{MMD} distances in order to align data marginal and conditional distributions between the source and target domain, and ignores discriminative knowledge within data. Here, a \textit{repulsive force}(\textbf{RF}) term $Dist_{{\cal S} \to {\cal T}}^{re} + Dist_{{\cal T} \to {\cal S}}^{re}$ is introduced to enable discriminative \textbf{DA} as shown in Fig.\ref{fig:div}.c. Specifically, we denote  ${{\cal S} \to {\cal T}}$ and ${{\cal T} \to {\cal S}}$ to index the distances computed from ${D_{\cal S}}$ to ${D_{\cal T}}$ and  ${D_{\cal T}}$ to ${D_{\cal S}}$, respectively, and $Dist_{{\cal S} \to {\cal T}}^{re}$ as the sum of the distances between each source sub-domain ${D_{\cal S}}^{(c)}$ and all the  target sub-domains ${D_{\cal T}}^{(r);\;r \in \{ \{ 1...C\}  - \{ c\} \} }$ excluding the $c$-th target sub-domain. Symmetrically, $Dist_{{\cal T} \to {\cal S}}^{re}$ is defined in a similar way as $Dist_{{\cal S} \to {\cal T}}^{re}$. These two distances are computed as:
\begin{equation}\label{eq:CDDAnew}
	\resizebox{0.9\hsize}{!}{%
		$\begin{array}{l}
		Dist_{S \to T}^{re} + Dist_{T \to S}^{re} = Dis{t^c}\sum\limits_{c = 1}^C {({D_S}^c,{D_T}^{r \in \{ \{ 1...C\}  - \{ c\} \} })} \\
		\;\;\;\;\;\;\;\;\;\;\;\;\;\;\;\;\;\;\;\;\;\;\;\;\;\;\;\;\;\;\;\;\; + Dis{t^c}\sum\limits_{c = 1}^C {({D_T}^c,{D_S}^{r \in \{ \{ 1...C\}  - \{ c\} \} })} \\
		\;\;\;\;\;\;\;\;\;\;\;\;\;\;\;\;\;\;\;\;\;\;\;\;\;\;\;\;\;\;\;\;\; = \sum\limits_{c = 1}^C {tr({{\bf{A}}^T}{\bf{X}}({{\bf{M}}_{S \to T}} + {{\bf{M}}_{T \to S}}){{\bf{X}}^{\bf{T}}}{\bf{A}})} 
		\end{array}$}
\end{equation}

Where:
\begin{itemize}
	\item {${{\bf{M}}_{{\cal S} \to {\cal T}}}$ is defined as: ${{{({{\bf{M}}_{{\bf{S}} \to {\bf{T}}}})}_{ij}} = \frac{1}{{n_s^{(c)}n_s^{(c)}}}}$ if $({{\bf{x}_i},{\bf{x}_j} \in {D_S}^{(c)}})$, ${  \frac{1}{{n_t^{(r)}n_t^{(r)}}}}$ if $({{\bf{x}_i},{\bf{x}_j} \in {D_T}^{(r)}})$, ${  \frac{{ - 1}}{{n_s^{(c)}n_t^{(r)}}}}$ if $({{\bf{x}_i} \in {D_S}^{(c)},{\bf{x}_j} \in {D_T}^{(r)}\;or\;{\bf{x}_i} \in {D_T}^{(r)},{\bf{x}_j} \in {D_S}^{(c)}})$ and ${  0}$ otherwise.}
	
	\item {${{\bf{M}}_{{\cal T} \to {\cal S}}}$ is defined as: ${{{({{\bf{M}}_{{\bf{T}} \to {\bf{S}}}})}_{ij}} = \frac{1}{{n_t^{(c)}n_t^{(c)}}}}$ if ${({\bf{x}_i},{\bf{x}_j} \in {D_T}^{(c)})}$, ${  \frac{1}{{n_s^{(r)}n_s^{(r)}}}}$ if ${({\bf{x}_i},{\bf{x}_j} \in {D_S}^{(r)})}$, ${  \frac{{ - 1}}{{n_t^{(c)}n_s^{(r)}}}}$ if ${({\bf{x}_i} \in {D_T}^{(c)},{\bf{x}_j} \in {D_S}^{(r)}\;or\;{\bf{x}_i} \in {D_S}^{(r)},{\bf{x}_j} \in {D_T}^{(c)})}$ and ${  0}$ otherwise.}
\end{itemize}

Therefore, maximizing Eq.(\ref{eq:CDDAnew}) increases the distances of each sub-domain with the other remaining sub-domains across domain, \textit{i.e.}, the between-class distances across domain, and thereby facilitates a discriminative \textbf{DA}. This across domain \textbf{RF} term was introduced in our previously proposed \textbf{DGA-DA} \cite{luo2020discriminative} and  has already shown its effectiveness.



\subsubsection{\textit{Repulsive force} term within the source domain}
\label{Repulsive force term across different domains}

While Sect.\ref{subsection:Matching Marginal and Conditional Distributions} and
sect.\ref{Repulsing interclass data for discriminative DA} have so far endeavored  to minimize the second term of the right-hand in Eq.(\ref{eq:bound}), we turn our attention here to optimize the first term of Eq.(\ref{eq:bound}) as shown in Fig.\ref{fig:div}.(d). Specifically, we introduce a \textit{repulsive force} term $Dist_{{\cal S} \to {\cal S}}^{re}$  (Fig.\ref{fig:div}.(d)), so as to increase the discriminative power on the labeled source domain,  thereby making it possible for a better predictive model on the source domain. Using ${{\cal S} \to {\cal S}}$ to index the distances computed from ${D_{\cal S}}$ to ${D_{\cal S}}$, we can compute, similarly as in eq.(\ref{eq:CDDAnew}),  $Dist_{{\cal S} \to {\cal S}}^{re}$ as the sum of the distances from each source sub-domain ${D_{\cal S}}^{(c)}$ to all the other source sub-domains ${D_{\cal S}}^{(r);\;r \in \{ \{ 1...C\}  - \{ c\} \} }$, excluding the $c$-th source sub-domain:

\begin{equation}\label{eq:stos}
	\resizebox{0.75\hsize}{!}{%
		$\begin{array}{l}
		Dist_{S \to S}^{re} = Dis{t^c}\sum\limits_{c = 1}^C {({D_S}^c,{D_S}^{r \in \{ \{ 1...C\}  - \{ c\} \} })} \\
		\;\;\;\;\;\;\;\;\;\;\;\;\;\;\; = \sum\limits_{c = 1}^C {tr({{\bf{A}}^T}{\bf{X}}({{\bf{M}}_{S \to S}}){{\bf{X}}^{\bf{T}}}{\bf{A}})} 
		\end{array}$}
\end{equation}
where ${{\bf{M}}_{{\cal S} \to {\cal S}}}$ is defined as: ${{{({{\bf{M}}_{{\bf{S}} \to {\bf{S}}}})}_{ij}} = \frac{1}{{n_s^{(c)}n_s^{(c)}}}}$ if ${({x_i},{x_j} \in {D_S}^{(c)})}$, ${ \frac{1}{{n_s^{(r)}n_s^{(r)}}}}$ if ${({x_i},{x_j} \in {D_S}^{(r)})}$, ${  \frac{{ - 1}}{{n_s^{(c)}n_s^{(r)}}}}$ if ${({x_i} \in {D_S}^{(c)},{x_j} \in {D_S}^{(r)}\;or\;{x_i} \in {D_S}^{(r)},{x_j} \in {D_S}^{(c)})}$ and ${ 0}$ otherwise.

Maximizing Eq.(\ref{eq:stos}) increases the between-class distances in the source domain and thereby optimizing the first term of the right-hand in Eq.(\ref{eq:bound}) on classification errors on the source domain. In our model, the \textbf{RF} term within the source domain as defined by Eq.(\ref{eq:stos}) added to the across domain \textbf{RF} term as defined by Eq.(\ref{eq:CDDAnew}) is named improved \textit{repulsive force} (\textbf{RF}) and optimized simultaneously.

\begin{figure}[h!]
	\centering
	\includegraphics[width=1\linewidth]{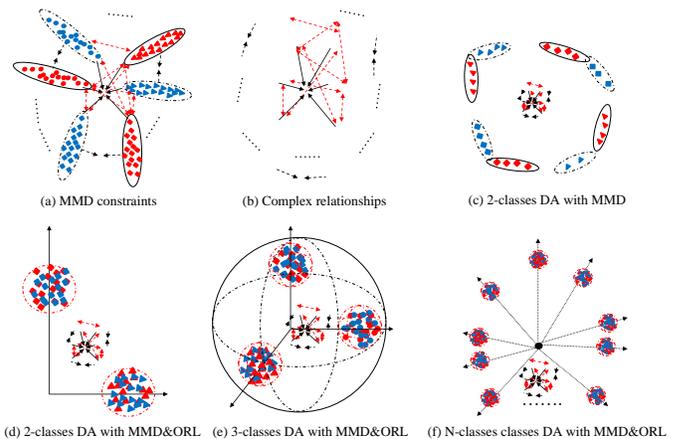}
	\caption{Illustration of MMD constraints and orthogonal projection based label consistent regression.} 
	\label{fig:2}
\end{figure}


\subsubsection{Noise Robust Sparse Orthogonal Label Regression}
\label{Label Consistent Regression}
The improved \textit{repulse force} term formulated in
sect.\ref{Repulsing interclass data for discriminative DA} and sect.\ref{Repulsive force term across different domains} aims to increase the between-class distances, across domain through Eq.(\ref{eq:CDDAnew}) and within the source domain through Eq.(\ref{eq:stos}),  and enable discriminative alignment of marginal and conditional data distributions (eq.(\ref{eq:JDA})) in sect.\ref{subsection:Matching Marginal and Conditional Distributions}), but does not seek to decrease the intra-class distances. As a result, situations as shown in Fig.\ref{fig:2}.(a,b,c), where the instances of a class are pushed away from other class instances but don't get closer each other within the class, could happen. More importantly, while eq.(\ref{eq:CDDAnew}) and eq.(\ref{eq:stos}) make it possible to decrease the classification errors of a hypothesis on the source domain, the first term of the error bound in eq.(\ref{eq:bound}) is not explicitly optimized. Here, we propose to solve the previous two issues through orthogonal label regression as shown in Fig.\ref{fig:2}.(d, e, f) where the instances of each class (sub-domain) across domain get close to their one-vs-all hot label vector which is orthogonal to those of other classes.

 To this end, we introduce  a novel \textit{orthogonal label regression} (\textbf{OLR}) constraint $\Phi ({\bf{A}},{{\bf{Y}}_S},{{\bf{X}}_S})$, where \textbf{A} is the transformation matrix projecting both the source and target data onto a latent shared feature subspace of dimension $k$. Specifically, we first introduce an embedding trick which consists of immersing a $C$-dimensional one-vs-all hot label vector into the $k$-dimensional latent shared feature subspace simply by adding $(k-C+1)$ times $0$, \textit{e.g.}, a $3$-dimensional one-vs-all label vector $(0,1,0)^t$ is represented by its corresponding one-vs-all hot vector $(0,1,0,0,0)^t$ in the $5$-dimensional feature subspace using the embedding trick. We can then perform least  square regression  (\textbf{LSR}): $\min \left\| {{{\bf{X}}^T}{\bf{A}} - {\bf{Y}}} \right\|_F^2$ ${\rm{s}}t.{\bf{Y}} \ge {\bf{0}},\;{\bf{Y1}} = {\bf{1}}$, with  \textbf{Y} the class label matrix as  defined in sect.\ref{Notations and Problem Statement} and extended into a $n \times k$ matrix by embedding each one-vs-all hot label vector into a $k$-dimensional one using the embedding trick. Minimization of \textbf{LSR} thus simply requires that each labeled sample be projected within the vicinity of its corresponding label hot vector in the $k$-dimensional feature space.   
 
 It is worth noting that the proposed \textbf{OLR} enjoys the following three properties: 
\begin{itemize}
\item \textbf{Orthogonality}, \textit{i.e.}, $({{\bf{Y}}_{{\rm{i}} = c}} \bullet {{\bf{Y}}_{{\rm{i}} \ne c}}) = 0$  with $\bullet$ denoting the dot product. This constraint simply expresses that one-vs-all hot label vectors in the shared latent feature subspace are orthogonal each other. As a result, projecting each data sample into the vicinity of its corresponding label vector keeps the samples of each sub-domain far away from those of other sub-domains and thereby improving data discriminativeness and optimizing term.1 of Eq.(\ref{eq:bound}); 
\item \textbf{Label Embedding Constraint}, \textit{i.e.}, $Q({{\mathbf{Y}}_{{\text{i}} \notin {\text{\{1}}...{\text{C\} }}}}|{\chi _S} \cup {\chi _T}) = 0$. This property simply denotes the fact that we have made use of the embedding trick for immersing each $C$ dimensional one-vs-all hot label vector into the $k$ dimensional shared feature space and there are no label vectors for classes ranging from $C+1$ to $k$; 
\item \textbf{Sharing of the feature projection and label regression matrix} through ${\mathbf{A}}$. Thanks to the label embedding constraint, the projection matrix ${\mathbf{A}}$ through eq.(\ref{eq:JDA}), eq.(\ref{eq:CDDAnew}) and eq.(\ref{eq:stos}) for the search of a shared latent feature space aligning discriminatively marginal and conditional data distributions between the source and target domain can be shared with the one used for the orthogonal label regression (\textbf{OLR}) constraint, thereby jointly optimizing term.1 and term.2 of Eq.(\ref{eq:bound}) within a single unified feature and label subspace. Furthermore, as shown in Fig.\ref{fig:2}.((d), (e)), data of each sub-domain, \textit{i.e.} data with a same label, from the source and target domain,  are projected within the vicinity of their corresponding one-vs-all hot label vector, thereby also decreasing Term.3 of Eq.(\ref{eq:bound}), \textit{i.e.}, the errors of their respective labelling functions on the source and target domain.     
\end{itemize}

However, data from the source and target domain can be noisy. We account for data noise through  an error matrix $\textbf{E}$ and the \textbf{OLR} constraint can  therefore be reformulated as: $\min \left\| {{{\bf{X}}^T}{\bf{A}} - {\bf{Y}} + {\bf{E}}} \right\|_F^2$  ${\rm{s}}t.\;{\bf{Y}} \ge {\bf{0}},\;{\bf{Y1}} = {\bf{1}}$. The error matrix $\textbf{E}$ makes possible a certain tolerance of errors when projecting data into the vicinity of its corresponding label vector in the latent shared feature space, thereby enabling to account for outliers and alleviating the influence of negative transfer.  Additionally, given the fact that, in real-life applications, \textit{e.g.} visual object recognition,  data of a given class generally lie within a manifold of much lower dimension in comparison with the original data space, \textit{e.g.}, pixel number of images, we further introduce a ${l_{2,1}}$-norm constraint so as to fulfill the property that the class label of a data sample  should be regressed from a sparse combination of features in the latent shared feature subspace. This constraint introduces a regularization term on \textbf{A} for discriminative subspace projection, which also  optimizes Term.3 of Eq.(\ref{eq:bound}). Putting all these together, the initial \textit{Orthogonal Label  Regression} (\textbf{OLR}) constraint becomes \textit{Noise Robust Sparse Orthogonal Label  Regression} (\textbf{NRS\_OLR}) and  is finally  formulated as:
\begin{equation}\label{eq:RLR}
	\resizebox{0.6\hsize}{!}{%
		$\begin{array}{*{20}{l}}
{\min \left\| {{{\bf{X}}^T}{\bf{A}} - {\bf{Y}} + 1{{\bf{e}}^T}} \right\|_F^2 + \beta \left\| {\bf{A}} \right\|_{2,1}^2}\\
{{\rm{s}}t.\;{\bf{Y}} \ge {\bf{0}},\;{\bf{Y1}} = {\bf{1}},({{\bf{Y}}_{{\rm{i}} = c}} \bullet {{\bf{Y}}_{{\rm{i}} \ne c}}) = 0}
\end{array}$}
\end{equation}


\subsubsection{The final model}
\label{The final model}
By integrating all the properties introduced in sect.\ref{subsection:Matching Marginal and Conditional Distributions} through
sect.\ref{Label Consistent Regression}, we obtain our final DA model, formulated as Eq.(\ref{eq:opti})

\begin{equation}\label{eq:opti}
	\resizebox{0.85\hsize}{!}{%
		$\begin{array}{*{20}{l}}
{\mathop {\min }\limits_{{\bf{A}},{\bf{e}},{{\bf{Y}}_{\bf{U}}} \ge 0,{{\bf{Y}}_{\bf{U}}}{\bf{1}} = {\bf{1}}} (tr({{\bf{A}}^T}{\bf{X}}{{\bf{M}}^*}{{\bf{X}}^T}{\bf{A}}) + \alpha \left\| {\bf{A}} \right\|_F^2 + \beta \left\| {\bf{A}} \right\|_{2,1}^2)}\\
{\;\;\;\;\;\;\;\;\;\;\;\;\;\;\;\;\;\;\;\;\;\;\;\;\;\; + \left\| {{{\bf{X}}^T}{\bf{A}} + {\bf{1}}{{\bf{e}}^T} - {\bf{Y}}} \right\|_F^2}\\
{{\rm{s}}t.{{\bf{M}}^*} = {{\bf{M}}_0} + \sum\limits_{c = 1}^C {({{\bf{M}}_c})}  - {{\bf{M}}_{REP}},\;{\bf{Y}} \ge {\bf{0}},\;{\bf{Y1}} = {\bf{1}},}\\
{\;\;\;\;\;{{\bf{A}}^T}{\bf{XH}}{{\bf{X}}^T}{\bf{A}} = {\bf{I}},({{\bf{Y}}_{{\rm{i}} = c}} \bullet {{\bf{Y}}_{{\rm{i}} \ne c}}) = 0}
\end{array}$}
\end{equation}


where ${{\bf{M}}_{REP}} = {{\bf{M}}_{S \to T}} + {{\bf{M}}_{T \to S}} + {{\bf{M}}_{S \to S}} $ is the improved overall \textit{repulsive force} constraint matrix, ${\bf{H}}{\rm{ = }}{\bf{I}} - \frac{1}{n}{\bf{1}}$ is the centering matrix, the constraint ${\bf{Y}} \ge {\bf{0}}$ and $\bf{Y1} = {\bf{1}}$ simply expresses the fact that each data sample has a label vector whose class probability sums to $1$,    whereas the constraint ${{{\bf{A}}^T}{\bf{XH}}{{\bf{X}}^T}{\bf{A}} = {\bf{I}}}$ derives from Principal  Component Analysis (\textbf{PCA}) to preserve the intrinsic data covariance of both domains and avoid the trivial solution for \textbf{A}. 

Through iterative optimization of Eq.(\ref{eq:opti}), our DA method searches jointly a well regularized discriminative latent feature subspace shared between the source and target domain and a noise robust sparse label regression model, thereby optimizing at the same time the three terms of the right-hand of Eq.(\ref{eq:bound}).

\subsection{Solving the model}
\label{Solving the model}
Eq.(\ref{eq:opti}) is not convex, therefore we propose an effective method which optimizes the key variables, \textit{e.g.}, ${\mathbf{A}},{\mathbf{M}},{\mathbf{Y}}$, in a coordinate descent manner. Main steps for solving Eq.(\ref{eq:opti}) are as follows. All the key steps have closed form solution:

\textit{\textbf{Step.1}} (Initialization of ${{\bf{M}}^*}$) ${{\bf{M}}^*}$ can be initialized by calculating ${{\bf{M}}_0}$ since there is no labels or pseudo labels on the target domain initially. We obtain ${{\bf{M}}^*} = {{\bf{M}}_0}$ where ${{\bf{M}}_0}$ is the MMD matrix as defined in Eq.(\ref{eq:JDA}).

\textit{\textbf{Step.2}} (Initialization of ${\bf{A}}$) Similar to \textbf{JDA}, ${\bf{A}}$ can be initialized to reduce the marginal and conditional distributions between $\mathcal{P}(\mathcal{X_S})$ and $\mathcal{P}(\mathcal{X_T})$ through an adaptive feature subspace  via the Rayleigh quotient algorithm, in solving Eq.(\ref{eq:S2}): 
\begin{equation}\label{eq:S2}
	\resizebox{0.7\hsize}{!}{%
		$({\bf{X}}({{\bf{M}}_{\bf{0}}} + \sum\nolimits_{{\rm{c}} = 1}^{c = C} {{{\bf{M}}_c}} ){{\bf{X}}^T} + \alpha {\bf{I}}){\bf{A}} = {\bf{XH}}{{\bf{X}}^T}{\bf{A}}\Phi$}
\end{equation}
where ${\bf{X}} = {{\bf{X}}_S} \cup {{\bf{X}}_T}$ and $\Phi  = diag({\phi _1},...,{\phi _k}) \in {R^{k \times k}}$ are Lagrange multipliers, $\sum\nolimits_{{\rm{c}} = 1}^{c = C} {{{\bf{M}}_c}} $ is obtained through labeled source domain data  ${{\bf{A}}^T}{{\bf{X}}_S}$ and  pseudo labels inferred on the target domain data ${{\bf{A}}^T}{{\bf{X}}_T}$. ${\bf{A}}$ is then initialized as the $k$ smallest eigenvectors of Eq.(\ref{eq:S2}), with $k$ defining the dimension of the latent shared feature subspace between the source and target domain.

\textit{\textbf{Step.3}} (Update of ${\bf{e}}$)  ${\bf{e}}$ is updated in solving Eq.(\ref{eq:opti}) with other variables held fixed. To update ${\bf{e}}$, one should solve  Eq.(\ref{eq:S41})	
\begin{equation}\label{eq:S41}
	\resizebox{0.53\hsize}{!}{%
		${\bf{e'}} = \arg \mathop {\min }\limits_{\bf{e}} \left\| {{{\bf{X}}^T}{\bf{A}} + {\bf{1}}{{\bf{e}}^T} - {\bf{Y}}} \right\|_F^2$}
\end{equation}

In setting to $0$ the partial derivative of Eq.(\ref{eq:S41}) with respect to $\textbf{e}$, we achieve the optimal solution of $\textbf{e}$ as	
\begin{equation}\label{eq:S42}
	\resizebox{0.4\hsize}{!}{%
		${\bf{e}} = \frac{1}{n}({{\bf{Y}}^T}{\bf{1}} - {{\bf{A}}^T}{\bf{X1}})$}
\end{equation}

\textit{\textbf{Step.4}} (Update of ${\bf{A}}$) ${\bf{A}}$ is updated by solving the optimization problem in Eq.(\ref{eq:opti}) with other variables held fixed. To ensure that Eq.(\ref{eq:opti}) is differentiable, we regularize $\left\| {\bf{A}} \right\|_{2,1}^2$ as $(\sum {_{j = 1}^k} \sqrt {\left\| {{{\bf{a}}^j}} \right\|_2^2 + \varepsilon } )$ to avoid $\left\| {\bf{A}} \right\|_{2,1}^2{\rm{ = }}{\bf{0}}$. As a result, Eq.(\ref{eq:opti}) becomes Eq.(\ref{eq:S51})
\begin{equation}\label{eq:S51}
	\resizebox{0.85\hsize}{!}{%
		$\begin{array}{l}
		{\bf{A'}} = \mathop {\arg \min }\limits_{{\bf{A}},{{\bf{A}}^T}{\bf{XH}}{{\bf{X}}^T}{\bf{A}} = {\bf{I}}} ((tr({{\bf{A}}^T}{\bf{X}}({{\bf{M}}^{\rm{*}}}){{\bf{X}}^T}{\bf{A}}) + \alpha \left\| {\bf{A}} \right\|_F^2\\
		+ \left\| {{{\bf{X}}^T}{\bf{A}} + {\bf{1}}{e^T} - {\bf{Y}}} \right\|_F^2{\rm{ + }}\beta (\sum\nolimits_{j = 1}^k {\sqrt {\left\| {{{\bf{a}}^j}} \right\|_2^2 + \varepsilon } } ))
		\end{array}$}
\end{equation}

$\varepsilon $ is infinitely close to zero, thereby making  Eq.(\ref{eq:S51}) closely equivalent to Eq.(\ref{eq:opti}).  Solving directly Eq.(\ref{eq:S51}) is non-trivial, we introduce a new variable ${\bf{G}} \in {R^{k*k}}$ which is a diagonal matrix with ${g_{jj}} = (\sum {_{i = 1}^k} \sqrt {\left\| {{{\bf{a}}^i}} \right\|_2^2 + \varepsilon } ) \div (\sqrt {\left\| {{{\bf{a}}^i}} \right\|_2^2 + \varepsilon } )$. $\textbf{G}$ and $\textbf{A}$ can be optimized iteratively. With  $\textbf{G}$ held fixed and  $\textbf{e}$  computed as in Eq.(\ref{eq:S42}), we can reformulate Eq.(\ref{eq:S51}) as Eq.(\ref{eq:S52})
\begin{equation}\label{eq:S52}
	\resizebox{0.89\hsize}{!}{%
		$\begin{array}{*{20}{l}}
		{{\bf{A'}} = \mathop {\arg \min }\limits_{{\bf{A}},{{\bf{A}}^T}{\bf{XH}}{{\bf{X}}^T}{\bf{A}} = {\bf{I}}} (tr({{\bf{A}}^T}{\bf{X}}({{\bf{M}}^{\rm{*}}}){{\bf{X}}^T}{\bf{A}}) + \alpha \left\| {\bf{A}} \right\|_F^2)}\\
		{\;\;\;\;\;\;\; + \left\| {{\bf{H}}{{\bf{X}}^T}{\bf{A}} - {\bf{HY}}} \right\|_F^2 + \beta tr({{\bf{A}}^T}{\bf{GA}})}
		\end{array}$}
\end{equation}


Eq.(\ref{eq:S52}) is a least square problem on the Stiefel manifold, which is a non-convex optimization problem. Therefore, it cannot be  directly solved via the Lagrangian method or an analytical solution. Inspired by previous research on solving quadratic problem on the Stiefel manifold\cite{fiori2005formulation,nie2014optimal,harandi2017dimensionality}, we propose a novel generalized power iteration (\textbf{GPI}) method to  optimize the projection matrix ${\bf{A}}$ that rotates the factor matrix to best fit the hypothesis subspace.

\textit{\textbf{Step.i}}
We propose  Cholesky factorization of ${\bf{H}}$, which aims to obtain a lower triangular matrix ${\bf{h}}$, so that ${\bf{h}}{{\bf{h}}^T} = {\bf{H}}$. 

\textit{\textbf{Step.ii}}
Eq.(\ref{eq:S52}) is reformulated as:
\begin{equation}\label{eq:qpsm}
	\resizebox{0.89\hsize}{!}{%
		$\begin{array}{l}
		\mathop {\arg \min }\limits_{{\bf{A}},{{\bf{A}}^T}{\bf{XH}}{{\bf{X}}^T}{\bf{A}} = {\bf{I}}} {\rm{tr(}}{{\bf{A}}^T}({\bf{X}}{{\bf{H}}^T}{\bf{H}}{{\bf{X}}^T} + \beta {\bf{G}} + \alpha {\bf{I}}\\
		\;\;\;\;\;\;\;\;\;\;\;\;\;\;\;\;\;\;\;\;\;\;\;\; + ({\bf{X}}({{\bf{M}}^{\rm{*}}}){{\bf{X}}^T})){\bf{A}}{\rm{) - 2tr(}}{{\bf{A}}^T}{\bf{X}}{{\bf{H}}^T}{\bf{HY}}{\rm{)}}
		\end{array}$}
\end{equation}
We set:
\begin{equation}\label{eq:wr}
	\resizebox{0.89\hsize}{!}{%
		$\begin{array}{*{20}{l}}
  {{{\mathbf{W}}^T} = {{\mathbf{A}}^T}{\mathbf{Xh}}} \\ 
  {{\mathbf{C}} = {{\mathbf{h}}^{ - 1}}{{\mathbf{H}}^T}{\mathbf{HY}}} \\ 
  {{\mathbf{B}} = {{({\mathbf{Xh}})}^{ - 1}}(({\mathbf{X}}{{\mathbf{H}}^T}{\mathbf{H}}{{\mathbf{X}}^T} + \beta {\mathbf{G}} + \alpha {\mathbf{I}} + ({\mathbf{X}}({{\mathbf{M}}^{\text{*}}}){{\mathbf{X}}^T}))){{({{\mathbf{h}}^T}{{\mathbf{X}}^T})}^{ - 1}}} 
\end{array}$}
\end{equation}
Using Eq.(\ref{eq:wr}), Eq.(\ref{eq:qpsm}) can be written for short as:
\begin{equation}\label{eq:qpsms}
	\resizebox{0.5\hsize}{!}{%
		$\mathop {\arg \min }\limits_{{\mathbf{W}},{{\mathbf{W}}^T}{\mathbf{W}} = {\mathbf{I}}} {\text{tr}}({{\mathbf{W}}^T}{\mathbf{BW}} - {\text{2}}{{\mathbf{W}}^T}{\mathbf{C}})$}
\end{equation}
where ${\bf{B}}$ is a symmetric matrix.

\textit{\textbf{Step.iii}}
Initialize ${{{\mathbf{W}}^T} = {{\mathbf{A}}^T}{\mathbf{Xh}}}$ to satisfy ${{\bf{W}}^T}{\bf{W}} = {\bf{I}}$, and set ${\bf{B'}} = \mu {\bf{I}} - {\bf{B}}$, which ensures ${\bf{B'}}$ is a positive definite matrix.

\textit{\textbf{Step.iv}}
Set ${\bf{Z}} = 2{\bf{B'W}} + 2{\bf{C}}$. Then, optimize ${\bf{US}}{{\bf{V}}^T} = {\bf{Z}}$ via singular value decomposition method on ${\bf{Z}}$.

\textit{\textbf{Step.v}}
Update ${{\bf{W}}^T} = {\bf{U}}{{\bf{V}}^T}$ and ${{\bf{A}}^T} = {{\bf{W}}^T}{({\bf{Xh}})^{ - 1}}$.

Eventually, the final optimization of Eq.(\ref{eq:S51}) is ${\bf{A'}} = {({{\bf{W}}^T}{({\bf{Xh}})^{ - 1}})^T}$. Algorithm 1 details the  whole process to update ${\bf{A'}}$.

\begin{algorithm}[!ht]
	\caption{Power iteration method for solving Eq.(\ref{eq:S51})}
	\KwIn{Data $\bf{X}$, Source domain label ${\bf{Y}}_{\cal S}$, MMD matrix ${{\bf{M}}^{\rm{*}}}$, fixed matrix ${\bf{G}}$, ${\bf{H}}$, regularization parameters $\beta $ and $\alpha $}
	\textbf{{1}}: Initialize ${\bf{W}}$ and ${\bf{B'}}$ as introduced in \textit{\textbf{Step.iii}};\\
	\textbf{2}: Update ${\bf{Z}},{{\bf{A}}^T},{{\bf{W}}^T}$;\\
	\eIf{ Non convergence }{
		(i) Do \textit{\textbf{Step.iv}}\;
		(ii) Do \textit{\textbf{Step.v}}\;
	}{
	\textbf{break}\;
}
\textbf{3}: Update ${{\bf{A'}}}$ via solving ${\bf{A'}} = {({\bf{W}}{({\bf{Xh}})^{ - 1}})^T}$\\
\KwOut{${\bf{A}} \leftarrow {\bf{A'}}$}
\end{algorithm}

\textit{\textbf{Step.5}} (Update of ${\bf{Y}}$)
The label matrix \textbf{Y} contains two  parts: true labels ${{\bf{Y}}_S} = {\{ {y_1},...,{y_{{n_s}}}\} ^T} \in {{\bf{\mathbb{R}}}^{{n_s} \times C}}$, and pseudo labels  ${{\bf{Y}}_T} = {\{ {y_{{n_s} + 1}},...,{y_{{n_s} + {n_t}}}\} ^T} \in {{\bf{\mathbb{R}}}^{{n_t} \times C}}$. Our aim is to iteratively refine the latter ones. Given fixed $\textbf{A}$, $\textbf{e}$ and ${{\bf{M}}^*}$,  each ${y_{\rm{i}}} \in {{\bf{Y}}_T}$ can be updated by solving the following problem:
\begin{equation}\label{eq:S61}
	\resizebox{0.56\hsize}{!}{%
		${y_i}^\prime  = \arg \mathop {\min }\limits_{{y_i} \ge 0,y_i^T{\bf{1}} = 1} \left\| {{{\bf{X}}^T}{\bf{A}} - {y_i}{\rm{ + }}{\bf{e}}} \right\|_F^2$}
\end{equation}

Using Lagrangian multipliers method, the final optimal solution of ${y_{\rm{i}}}$ is 
\begin{equation}\label{eq:S62}
	\resizebox{0.35\hsize}{!}{%
		${y_i}^\prime  = ({{\bf{A}}^T}{{\bf{x}}_i} + e + \partial )$}
\end{equation}
where $\partial $ is coefficient of Lagrangian constraint $y_{\rm{i}}^T{\bf{1}} - 1 = 0$, which can be obtained by solving $y_{\rm{i}}^T{\bf{1}} = 1$.

\textit{\textbf{Step.6}} (Update of ${{\bf{M}}^*}$)
With the labeled source domain data  ${{\bf{A}}^T}{{\bf{X}}_S}$ and  the labels inferred on the target domain data ${{\bf{A}}^T}{{\bf{X}}_T}$ as in \textbf{Step.5}, we can update  ${{\bf{M}}^*}$ as
\begin{equation}\label{eq:S3}
	\resizebox{0.5\hsize}{!}{%
		${{\bf{M}}^*} = {{\bf{M}}_0} + \sum _{c = 1}^C({{\bf{M}}_c}) - {{\bf{M}}_{REP}}$}
\end{equation}	
where ${{{\bf{M}}_c}}$ and ${{{\bf{M}}_{REP}}}$ are defined in Eq.(\ref{eq:opti}).	

The complete learning algorithm is summarized in Algorithm 2 - \textbf{DOLL-DA}.


\begin{algorithm}[!ht]
	\caption{Discriminative  Label  Consistent Domain Adaptation (DOLL-DA)}
	\KwIn{Data $\bf{X}$, Source domain label ${\bf{Y}}_{\cal S}$, subspace dimension $k$, iterations $T$, regularization parameters $\beta $ and $\alpha $}
	\textbf{{1}}: Initialize ${{\bf{M}}^*} = {{\bf{M}}_0}$ as defined in Eq.(\ref{eq:JDA}) ;\\
	\textbf{2}: Initialize $\textbf{A}$ by solving Eq.(\ref{eq:S2}); ($t: = 0$) \\
	\While{$\sim isempty(\bf{X},{{\bf{Y}}_{\cal S}})$ and $t<T$	}{
		\textbf{3}: Update ${{\bf{M}}^*}$ by solving Eq.(\ref{eq:S3}) \\
		\textbf{4}: Update ${\bf{e}}$ by solving Eq.(\ref{eq:S42})  \\
		\textbf{5}: Update ${\bf{A}}$; (${t_1}: = 0.$)\\
		\eIf{ ${t_1} < T$ }{
			(i) Initialize ${\bf{G}}$ as an identity matrix\;
			(ii) Update ${\bf{A}}$ by solving ${\bf{Algorithm 1}}$\;
			(iii) Update ${\bf{G}}$ by calculating ${g_{jj}} = (\sum {_{i = 1}^k} \sqrt {\left\| {{{\bf{a}}^i}} \right\|_2^2 + \varepsilon } ) \div (\sqrt {\left\| {{{\bf{a}}^i}} \right\|_2^2 + \varepsilon } )$\;
			(iv) ${t_1} = {t_1} + 1$\;
		}{
		\textbf{break}\;
	}
	\textbf{6}: Update ${\bf{Y}}$ by solving Eq.(\ref{eq:S62})\\
	\textbf{7}: Update pseudo target labels $ {\bf{Y}}_{\cal T}^{(T)} = {{\bf{Y}}}\left[ {:,({n_s} + 1):({n_s} + {n_t})} \right] $;\\
	\textbf{8}:$t=(t+1)$;\\               
	
}
\KwOut{${\bf{A}}$, ${\bf{Z}} = {{\bf{A}}^T}{\bf{X}}$, $\textbf{Y}$}
\end{algorithm}


\subsection{Kernelization}
\label{Kernelization Analysis}
The proposed \textbf{DOLL-DA} method is extended to nonlinear problems in a Reproducing Kernel Hilbert Space via the kernel mapping $\phi :x \to \phi (x)$, or $\phi ({\bf{X}}):[\phi ({{\bf{x}}_1}),...,\phi ({{\bf{x}}_n})]$, and the kernel matrix ${\bf{K}} = \phi {({\bf{X}})^T}\phi ({\bf{X}}) \in {R^{n*n}}$. We utilize
the representer theorem to formulate the Kernel \textbf{DOLL-DA} as:

\begin{equation}\label{eq:opti}
	\resizebox{0.85\hsize}{!}{%
		$\begin{array}{*{20}{l}}
{\mathop {\min }\limits_{{\bf{A}},{\bf{e}},{{\bf{Y}}_{\bf{U}}} \ge 0,{{\bf{Y}}_{\bf{U}}}{\bf{1}} = {\bf{1}}} (tr({{\bf{A}}^T}{\bf{K}}{{\bf{M}}^*}{{\bf{K}}^T}{\bf{A}}) + \alpha \left\| {\bf{A}} \right\|_F^2 + \beta \left\| {\bf{A}} \right\|_{2,1}^2)}\\
{\;\;\;\;\;\;\;\;\;\;\;\;\;\;\;\;\;\;\;\;\;\;\;\;\;\; + \left\| {{{\bf{K}}^T}{\bf{A}} + {\bf{1}}{{\bf{e}}^T} - {\bf{Y}}} \right\|_F^2}\\
{{\rm{s}}t.{{\bf{M}}^*} = {{\bf{M}}_0} + \sum\limits_{c = 1}^C {({{\bf{M}}_c})}  - {{\bf{M}}_{REP}},\;{\bf{Y}} \ge {\bf{0}},\;{\bf{Y1}} = {\bf{1}},}\\
{\;\;\;\;\;{{\bf{A}}^T}{\bf{KH}}{{\bf{K}}^T}{\bf{A}} = {\bf{I}},({{\bf{Y}}_{{\rm{i}} = c}} \bullet {{\bf{Y}}_{{\rm{i}} \ne c}}) = 0}
\end{array}$}
\end{equation}

\subsection{Time Complexity Analysis}
\label{subsection:Time Complexity Analysis}
Given $n$ the number of data samples including both the source and target domain and $l$ the feature dimension,  we denote by  $t$, ${t_1}$ the number of iterations with $t,{t_1} \prec \min (l,n)$. 	The major computational burden of the proposed Algorithm.2 lies in Step 2, 3 and 5 as sketched in Sect.\ref{Solving the model}. In Step 2, the singular value decomposition(SVD) is computed on a $n*n$ matrix, its computational complexity is $O({n^3})$. Step 3. constructs the  ${{\bf{M}}_{{\bf{cyd}}}}$ matrix, whose computational complexity is $O(4C{n^2})$ with $C$ the number of classes, \textit{i.e.}, the class cardinality.  Step 4.(ii) makes use of SVD to solve optimization on a $n*k$ size matrix as introduced in step.(iv) of Algorithm.1, its computational complexity is thus $O({n^2}k + n{k^2} + {k^3})$. In Step.5 and Step.6, $O({nk})$ operations are required for all other lines. Therefore, the overall computational complexity of the proposed Algorithm - \textbf{DOLL-DA}  is $O({n^3} + 4tC{n^2} + tnk + t{t_1}({n^2}k + n{k^2} + {k^3}))$.


\section{Experiments}
\label{Experiments}


\subsection{Benchmarks and Features}
\label{subsection:Benchmarks and Features}
\begin{figure}[h!]
	\centering
	\includegraphics[width=1\linewidth]{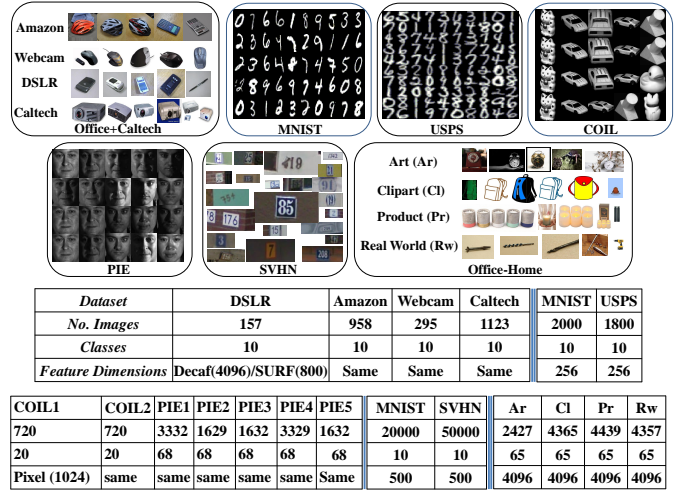}
	\caption { Sample images from eight datasets used in our experiments. Each dataset represents a different domain. The Office dataset in Office+Caltech contains three sub-datasets, namely DSLR, Amazon and Webcam.} 
	\label{fig:data}	
\end{figure}

As illustrated in Fig.\ref{fig:data}, USPS\cite{DBLP:journals/pami/Hull94}+MINIST\cite{lecun1998gradient}, COIL20\cite{long2013transfer}, PIE\cite{long2013transfer}, office+Caltech\cite{long2013transfer}, Office-Home\cite{venkateswara2017deep} and SVHN-MNIST\cite{bousmalis2016domain} are standard benchmarks for the purpose of evaluation and comparison with state-of-the-art in DA. In this paper, we follow the data preparation as most previous works\cite{DBLP:journals/pami/GhifaryBKZ17,DBLP:journals/tip/DingF17,DBLP:journals/corr/LuoWHC17,bousmalis2016domain,liang2018aggregating} do. We construct 49 datasets for different image classification tasks.

\textbf{Office+Caltech} consists of 2533 images of 10 categories (8 to 151 images per category per domain)\cite{DBLP:journals/pami/GhifaryBKZ17}. These images come from four domains: (A) AMAZON, (D) DSLR, (W) WEBCAM, and (C) CALTECH. AMAZON images were acquired in a controlled environment with studio lighting. DSLR consists of high resolution images captured by a digital SLR camera in a home environment under natural lighting. WEBCAM images were acquired in a similar environment to DSLR, but with a low-resolution webcam. CALTECH images were collected from Google Images. 

We use two types of image features extracted from these datasets, \textit{i.e}.,   \textbf{SURF} and \textbf{DeCAF6}, that are publicly available.  The \textbf{SURF}\cite{gong2012geodesic} features are \textit{shallow} features extracted and quantized into an 800-bin histogram using a codebook computed with K-means on a subset of images from  Amazon. The resultant histograms are further standardized by z-score. The \textbf{Deep Convolutional Activation Features (DeCAF6)}\cite{DBLP:conf/icml/DonahueJVHZTD14} are \textit{deep} features computed as in \textbf{AELM}\cite{DBLP:journals/tcyb/UzairM17} which makes use of VLFeat MatConvNet library with different pretrained CNN models, including in particular  the Caffe implementation of \textbf{AlexNet}\cite{krizhevsky2012imagenet} trained on the ImageNet dataset. The outputs from the 6th layer are used as \textit{deep} features, leading to 4096 dimensional \textbf{DeCAF6} features. In this experiment, we denote the dataset \textbf{Amazon}, \textbf{Webcam}, \textbf{DSLR}, and \textbf{Caltech-256} as \textbf{A}, \textbf{W}, \textbf{D}, and \textbf{C}, respectively. The arrow “$\rightarrow$” is proposed to denote the direction from “source” to “target”. For example, “W $\rightarrow$ D” means the Webcam image dataset is considered as the labeled \textit{source} domain whereas the DSLR image dataset the unlabeled \textit{target} domain.

\textbf{USPS+MNIST} shares 10 common digit categories from two subsets, namely USPS and MNIST, but with very different data distributions (see Fig.\ref{fig:data}). We construct a first DA task \emph{USPS vs MNIST} by randomly sampling first 1,800 images in USPS to form the source data, and then 2,000 images in MNIST to form the target data. Then, we switch the source/target pair to get another DA task, \textit{i.e.}, \emph{MNIST vs USPS}. We uniformly rescale all images to size $16 \times 16$, and represent each one by a feature vector encoding the gray-scale pixel values. We also extract deep feature from softmax layer\cite{rozantsev2018beyond} of LeNet\cite{lecun1998gradient} architecture, leading to a 10 dimensional feature. Thus the source and target domain data share the same feature space. As a result, we have defined two cross-domain DA tasks, namely \emph{USPS $\rightarrow$ MNIST} and \emph{MNIST $\rightarrow$ USPS}.

\textbf{COIL20} contains 20 objects with 1440 images (Fig.\ref{fig:data}). The images of each object were taken in varying its pose about 5 degrees, resulting in 72 poses per object. Each image has a resolution of 32×32 pixels and 256 gray levels per pixel. In this experiment, we partition the dataset into two subsets, namely COIL 1 and COIL 2\cite{DBLP:journals/tip/XuFWLZ16}. COIL 1 contains all images taken within the directions in $[{0^0},{85^0}] \cup [{180^0},{265^0}]$ (quadrants 1 and 3), resulting in 720 images. COIL 2 contains all images taken in the directions
within $[{90^0},{175^0}] \cup [{270^0},{355^0}]$ (quadrants 2 and 4) and thus the number of images is also 720. In this way, we construct two subsets with relatively different distributions. In this experiment, the COIL20 dataset with 20 classes is split into two DA tasks, \textit{i.e.},  \emph{ COIL1 $\rightarrow$ COIL2} and \emph{COIL2 $\rightarrow$ COIL1}.

\textbf{PIE} face database consists of 68 subjects with each under 21 various illumination conditions\cite{DBLP:journals/tip/DingF17,long2013transfer}. We adopt five pose subsets: C05, C07, C09, C27, C29, which provide a rich basis for domain adaptation, that is, we can choose one pose as the source and any remaining one as the target. Therefore, we obtain $5 \times 4=20$ different source/target combinations. Finally, we combine all five poses together to form a single dataset for large-scale transfer learning experiment. We crop all images to $32 \times 32$ and only adopt the pixel values as the input. Finally, with different face {poses}, of which five subsets are selected, denoted as PIE1, PIE2, \textit{etc}., resulting in $5 \times 4=20$ DA tasks, \textit{i.e.}, \emph{PIE1 vs PIE 2} $\dots$ \emph{PIE5 vs PIE 4}, respectively.

\textbf{Office-Home} dataset as shown in Fig.\ref{fig:data} is a novel DA dataset recently introduced in \cite{venkateswara2017deep}. This dataset contains 4 domains. Each domain contains 65 categories. This dataset is used in a similar manner as the \textbf{Office+Caltech} dataset.  From the 4 domains, \textit{i.e.,} the Art (Ar), Clipart (Cl), Product (Pr) and Real-World (Rw), we generate 12 DA tasks, namely namely \emph{Ar} $\rightarrow$ \emph{Cl} $\dots$ \emph{Pr} $\rightarrow$ \emph{Rw}, respectively. \textbf{DeCAF6} features are extracted to evaluate the performance of the proposed DA algorithms on this dataset.

\textbf{SVHN-MNIST} contains the MNIST dataset  as introduced in \textbf{USPS-MNIST} but also the Street View House Numbers (SVHN) which is a collection of house numbers collected from Google street view images (see Fig.\ref{fig:data}). SVHN is quite distinct from the dataset of handwriting digits, \textit{i.e.}, digits in MNIST. Moreover, all the 2 domains are quite large, each having at least 60k samples over 10 classes. We propose to make use of the LeNet architecture \cite{lecun1998gradient}  and domain classifier as introduced in \cite{rozantsev2018beyond} to extract features for our DA tasks.

\subsection{Baseline Methods}
\label{subsection:Baseline Methods}
The proposed \textbf{DOLL-DA} method is compared with \textbf{thirty-two} methods of the literature, including deep learning-based approaches for unsupervised domain adaption. They are:

\begin{itemize}
	\item \textbf{Shallow methods}:
(1) 1-Nearest Neighbor Classifier(\textbf{NN}); 
(2) Principal Component Analysis (\textbf{PCA}); 
(3) \textbf{GFK} \cite{gong2012geodesic}; 
(4) \textbf{TCA} \cite{pan2011domain}; 
(5) \textbf{TSL} \cite{4967588}; 
(6) \textbf{JDA} \cite{long2013transfer}; 
(7) \textbf{ELM} \cite{DBLP:journals/tcyb/UzairM17}; 
(8) \textbf{AELM} \cite{DBLP:journals/tcyb/UzairM17}; 
(9) \textbf{SA} \cite{DBLP:conf/iccv/FernandoHST13}; 
(10) \textbf{mSDA} \cite{DBLP:journals/corr/abs-1206-4683}; 
(11) \textbf{TJM} \cite{DBLP:conf/cvpr/LongWDSY14}; 
(12) \textbf{RTML} \cite{DBLP:journals/tip/DingF17}; 
(13) \textbf{SCA} \cite{DBLP:journals/pami/GhifaryBKZ17}; 
(14) \textbf{CDML} \cite{DBLP:conf/aaai/WangWZX14}; 
(15) \textbf{LTSL} \cite{DBLP:journals/ijcv/ShaoKF14}; 
(16) \textbf{LRSR} \cite{DBLP:journals/tip/XuFWLZ16}; 
(17) \textbf{KPCA} \cite{DBLP:journals/neco/ScholkopfSM98}; 
(18) \textbf{JGSA}  \cite{Zhang_2017_CVPR}; 
(19) \textbf{CORAL}  \cite{sun2016return};
(20) \textbf{RVDLR}  \cite{jhuo2012robust};
(21) \textbf{LPJT}  \cite{li2019locality};
(22) \textbf{DGA-DA}\cite{luo2020discriminative}.

\item \textbf{Deep methods}: 
(23) \textbf{AlexNet}  \cite{krizhevsky2012imagenet};
(24) \textbf{DAH}  \cite{venkateswara2017deep}; 
(25) \textbf{DANN}  \cite{ganin2016domain};
(26) \textbf{ADDA} \cite{tzeng2017adversarial};
(27) \textbf{LTRU}) \cite{sener2016learning};
(28) \textbf{ATU} \cite{DBLP:conf/icml/SaitoUH17};
(29) \textbf{BSWD} \cite{rozantsev2018beyond};
(30) \textbf{DSN} \cite{bousmalis2016domain};	
(31) \textbf{DDC} \cite{DBLP:journals/corr/TzengHZSD14};
(32) \textbf{DAN}  \cite{long2015learning}.
\end{itemize}

Direct comparison of the proposed \textbf{DOLL-DA} using shallow features against these \textbf{DL}-based DA approaches could be unfair. However, in order to give an idea on the performance gap between shallow and deep DA methods, we still compare their results with those of our shallow \textbf{DOLL-DA}.
For the purpose of fair comparison, we follow   the experiment settings of \textbf{DGA-DA}, \textbf{JGSA} and \textbf{BSWD}, and apply DeCAF6 as the input features for some methods to be evaluated. Whenever possible,  the reported performance scores of the \textbf{thirty-two} methods of the literature are directly  collected from their original papers or previous research  \cite{tzeng2017adversarial,DBLP:journals/tcyb/UzairM17,li2019locality,DBLP:journals/pami/GhifaryBKZ17,rozantsev2018beyond,Zhang_2017_CVPR,luo2020discriminative}. They are assumed to be their \emph{best} performance.

\subsection{Experimental Setup}
\label{subsection: Experimental setup}
For the problem of domain adaptation, it is not possible to tune a set of optimal hyper-parameters, given the fact that the target domain has no labeled data. Following the setting of previous research\cite{luo2020discriminative,long2013transfer,DBLP:journals/tip/XuFWLZ16}, we also evaluate the proposed \textbf{DOLL-DA} by empirically searching in the parameter space for the \emph{optimal} settings. Specifically, the proposed \textbf{DOLL-DA}  method has three hyper-parameters, \textit{i.e.}, the subspace dimension $k$, regularization parameters $\beta $ and $\alpha $. In  our experiments, we set $k = 300$ and 1) $\beta  = 0.1$, and $\alpha  = 1$ for \textbf{USPS}, \textbf{MNIST}， \textbf{COIL20} and \textbf{PIE}, 2) $\beta  = 1$, $\alpha  = 1$ for \textbf{Office+Caltech}, \textbf{SVHN-MNIST} and \textbf{Office-Home}.

In our experiment, {\emph{accuracy}}  on the test dataset as defined by Eq.(\ref{eq:accuracy}) is the performance measurement. It is widely used in literature, \textit{e.g.},\cite{long2015learning,DBLP:journals/corr/LuoWHC17,long2013transfer,DBLP:journals/tip/XuFWLZ16}, \textit{etc}.

\begin{equation}\label{eq:accuracy}
	\begin{array}{c}
		Accuracy = \frac{{\left| {x:x \in {D_T} \wedge \hat y(x) = y(x)} \right|}}{{\left| {x:x \in {D_T}} \right|}}
	\end{array}
\end{equation}
where ${\cal{D_T}}$ is the target domain treated as test data, ${\hat{y}(x)}$ is the predicted label and ${y(x)}$ is the ground truth label for the test data  $x$.

 The core model of the proposed \textbf{DOLL-DA} method is built on \textbf{JDA}, but adds up two optimization terms, namely discriminative data distribution alignment (\textbf{DDA}) term as defined in Eq.(\ref{eq:CDDAnew}, \ref{eq:stos}), and Noise Robust Sparse Orthogonal Label  Regression (\textbf{NRS\_OLR}) term as defined in Eq.(\ref{eq:RLR}). For further insight into the proposed \textbf{DOLL-DA} and the rational \textit{w.r.t} the \textbf{DDA} and \textbf{NRS\_OLR} term, respectively, we derive from Eq.(\ref{eq:opti}) three additional partial models, namely \textbf{OLR}, \textbf{CDDA+} and \textbf{JOLR-DA}:


\begin{itemize}
	
	\item \textbf{OLR}: In this setting, the partial model only makes use of the \textbf{NRS\_OLR} term, \textit{i.e.}, noise robust sparse orthogonal label regression term as in Eq.(\ref{eq:RLR})  and ignores the rest of our final model, \textit{i.e.}, data distribution alignment term as defined in Eq.(\ref{eq:JDA}) as well as  \textit{Discriminative MMD constraint} terms as defined in Eq.(\ref{eq:CDDAnew}, \ref{eq:stos}). This partial model amounts to make use of a particular classifier, \textit{i.e.},  noise robust sparse orthogonal label regression, and enables to quantify the importance of data distribution alignment in DA when contrasted with the baseline \textbf{JDA}; 

	\item \textbf{CDDA+}: In this setting,  the regularization term of \textbf{NRS\_OLR} (Eq.(\ref{eq:RLR})) is simply replaced by the Nearest Neighbor (NN) predictor.  This correspond to our final DA model as defined in Eq.(\ref{eq:S2}) and Eq.(\ref{eq:S3}) which only make use of the \textit{Discriminative MMD constraint} terms but without the \textbf{NRS\_OLR}  term as defined by Eq.(\ref{eq:RLR}). In comparison with \textbf{CDDA}, \textit{i.e.}, Close yet Discriminative DA, the partial model already studied in our former \textbf{DGA-DA} method \cite{luo2020discriminative} which demonstrated  the effectiveness of \textit{discriminative force}(\textbf{RF}) term in \textbf{DA}, \textbf{CDDA+} includes the improved  \textbf{RF} term which adds the \textbf{RF}  within the source domain as defined by Eq.(\ref{eq:stos}) to the across domain \textbf{RF} defined in Eq.(\ref{eq:CDDAnew}) as in \textbf{CDDA}. This partial model makes it possible to emphasize the interest of the joint optimization of the improved \textbf{RF} and \textbf{NRS\_OLR} terms in contrasting \textbf{DOLL-DA} with \textbf{CDDA+}; 


	\item \textbf{JOLR-DA}: In this setting, the partial model of the proposed method only cares about data distribution alignment as in \textbf{JDA} as well as the newly introduced \textbf{NRS\_OLR} term but ignores the \textit{discriminative force} as defined in Sect.\ref{Repulsing interclass data for discriminative DA} and Sect.\ref{Repulsive force term across different domains}. In studying \textbf{JOLR-DA}, we aim to highlight:  1)  the contribution of \textbf{NRS\_OLR} in regularizing the MMD constraints in contrasting \textbf{DOLL-DA} with \textbf{JOLR-DA}; 2) the effectiveness of \textbf{NRS\_OLR} and the proposed joint optimization strategy, in confronting \textbf{JOLR-DA} with \textbf{JDA}.



	
	\item \textbf{DOLL-DA}: This setting correspond to our full final model as defined in Eq.(\ref{eq:opti}). It thus contains both \textbf{CDDA+} as defined in Eq.(\ref{eq:JDA}, \ref{eq:CDDAnew}, \ref{eq:stos}) 
 and the \textbf{NRS\_OLR} term as defined by  Eq.(\ref{eq:RLR}) in sect. \ref{Label Consistent Regression}. 
	
\end{itemize}

\subsection{Experimental Results and Discussion}
\label{subsection: Experimental Results and Discussion}

\subsubsection{\textbf{Experiments on the CMU PIE Data Set}}
\label{subsubsection:Experiments on the CMU PIE dataset}

\begin{figure*}[h!]
	\centering
	\includegraphics[width=0.8\linewidth]{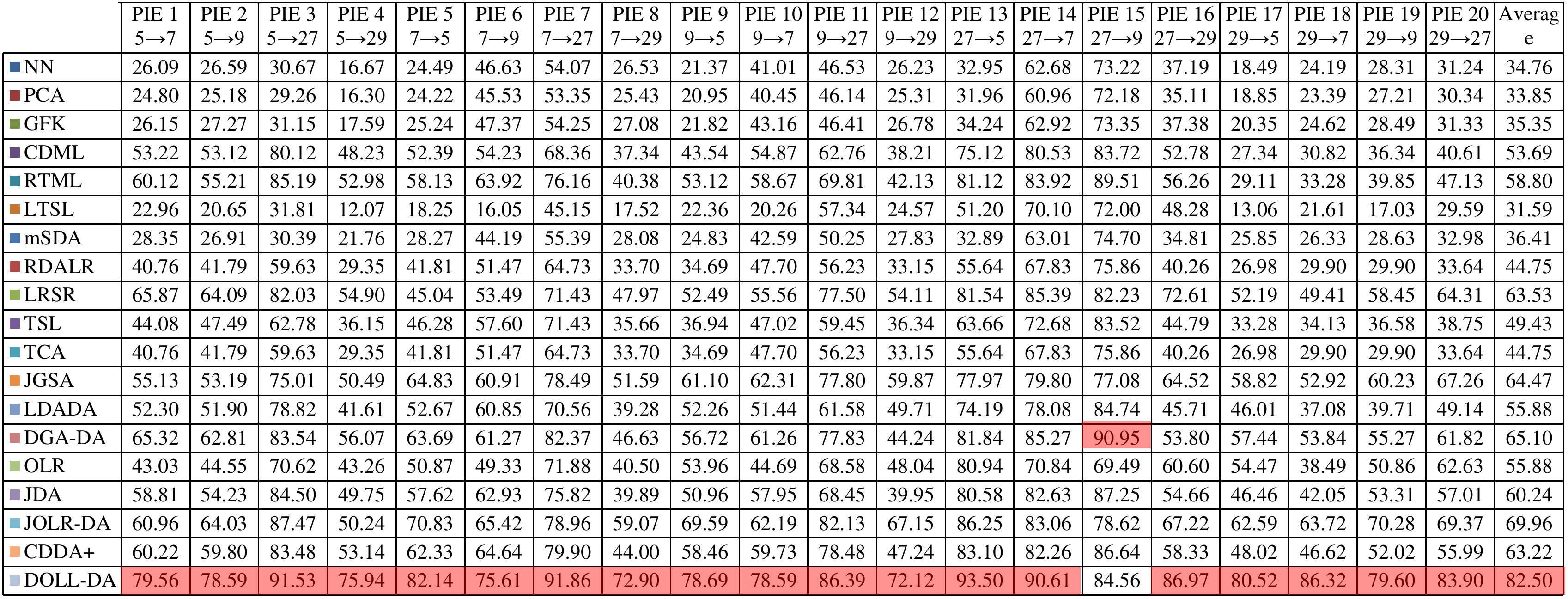}
	\caption { Accuracy${\rm{\% }}$ on the PIE Images Dataset.} 
	\label{fig:accPIE}
\end{figure*}

The CMU PIE dataset is a large face dataset featuring both illumination and pose variations. Fig.\ref{fig:accPIE} synthesizes the experimental results for \textbf{DA} using this dataset, where top results are highlighted in red color. As expected, without data distribution alignment,\textbf{OLR}, with $55.88\%$ average accuracy, performs worse than the base line \textbf{JDA} with $60.24\%$ average accuracy. In accounting for the discriminative force in data distribution alignment, \textbf{CDDA+} improves over \textbf{JDA} by 3 points and achieves $63.22\%$ average accuracy. In adding noise robust sparse orthogonal label regression to \textbf{JDA}, \textbf{JOLR-DA} achieves $69.96\%$ average accuracy and improves over \textbf{JDA} by 9 points, thereby demonstrating the effectiveness of the \textbf{NRS\_OLR} term. Now our final model, \textbf{DOLL-DA}, with $82.50\%$ average accuracy, achieves the state of the art performance on this dataset and improves over the baseline \textbf{JDA} by 22 points and the former state of the art DA method, \textit{i.e.}, \textbf{DGA-DA}, by a large margin of 17 points, thereby demonstrates with force the interest of joint optimization of discriminative data distribution terms and the \textbf{NRS\_OLR} term.

\subsubsection{\textbf{Experiments on the COIL 20 Dataset}} 
\label{subsubsection: results on the COIL dataset}
\begin{figure}[h!]
	\centering
	\includegraphics[width=1\linewidth]{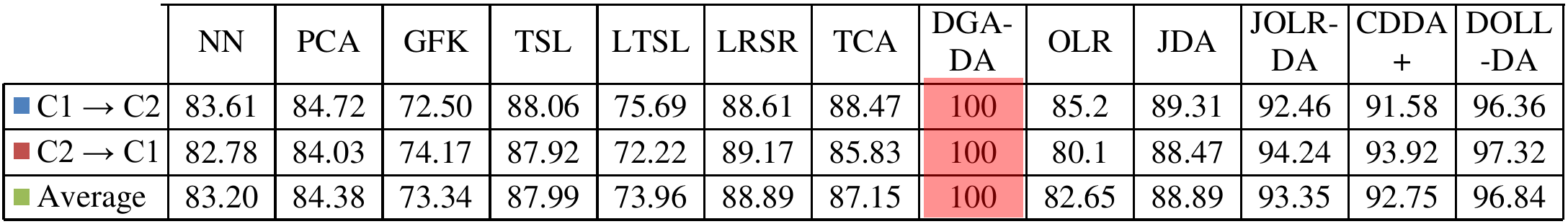}
	\caption { Accuracy${\rm{\% }}$ on the COIL Images Dataset.} 
	\label{fig:accCOIL}
\end{figure}

The \textbf{COIL} dataset (see fig.\ref{fig:data}) features the challenge of pose variations between the source and target domain. Fig.\ref{fig:accCOIL} reports the experimental results on the COIL dataset and displays similar patterns as those on the PIE dataset. \textbf{OLR} performs worse than \textbf{JDA} which is improved by \textbf{CDDA+} and \textbf{JOLR-DA}. Finally, \textbf{DOLL-DA} achieves $96.84\%$ and further improves \textbf{CDDA+} and \textbf{JOLR-DA} by 4 and 3 points, respectively. It is interesting to note that \textbf{DGA-DA} achieves $100\%$ average accuracy on this dataset and thereby outperforms \textbf{DOLL-DA} by $3.16$ points. However, \textbf{DGA-DA}'s outstanding performance is mainly related to the two particularities of the COIL 20 dataset. Indeed, the COIL dataset synthesizes 20 objects as foreground in varying their pose while the background is \textit{pure black}, therefore each sub-domain contains a single object and is naturally distributed within a specific manifold. Furthermore, the COIL 20 dataset merely contains 20 classes in contrast with 68 classes in the PIE dataset, thereby its classes are much more separated than those in PIE. As a matter of fact, the most simple baseline \textbf{NN}, \textit{i.e.}, Nearest Neighbor, already achieves a high average accuracy of $83.20\%$  on COIL 20  while it only displays $34.76\%$ average accuracy on PIE. As a result, in explicitly modeling the hidden data manifold structure through a Laplace graph, \textbf{DGA-DA} performs better label inference than \textbf{DOLL-DA}. 

\subsubsection{\textbf{Experiments on the Office-Home Dataset}}
\label{subsubsection:Experiments on the Office-Home dataset}

\begin{figure}[h!]
	\centering
	\includegraphics[width=0.9\linewidth]{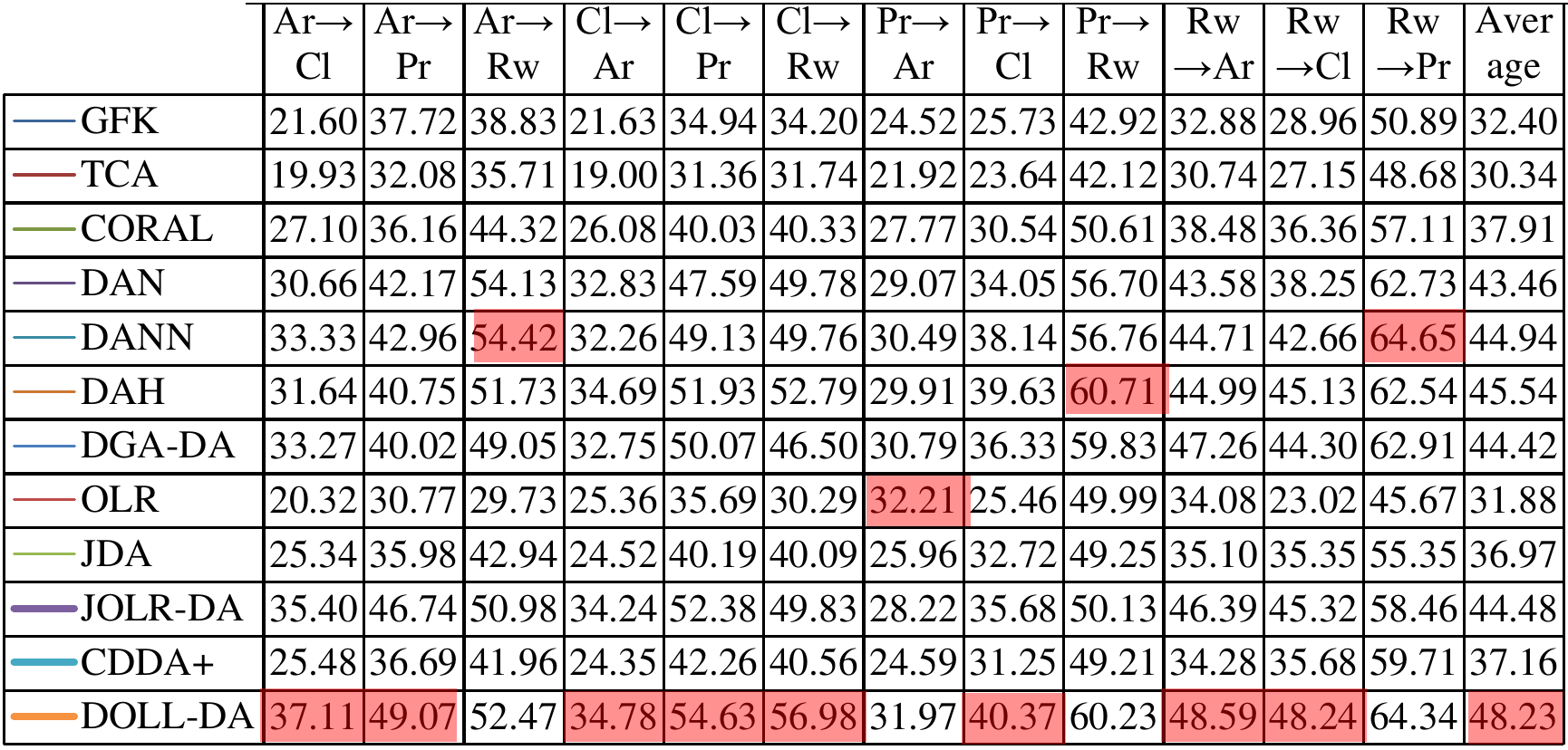}
	\caption { Accuracy${\rm{\% }}$ on the Office-Home Images Dataset.} 
	\label{fig:accHOME}
\end{figure}

As introduced in \textbf{DAH}\cite{venkateswara2017deep}, \textbf{Office-Home}  is a novel challenging benchmark for the DA task. It contains 4 very different domains with 65 object categories, thereby generating 12 different DA tasks. Fig.\ref{fig:accHOME} synthesizes the performance of the proposed DA methods with DeCAF6 features in comparison with the state of the art methods. Both \textbf{DAH} and \textbf{DAN} are deep DA methods and make use of multi-kernel \textbf{MMD} in aligning the source and target domain distributions. With $43.46\%$ and $45.54\%$ average accuracy, respectively, they surpass \textbf{JDA} with a margin up to 8 points. In extending \textbf{JDA} with repulsive force terms, \textbf{CDDA+} slightly improves \textbf{JDA} by $0.19$ point. However, thanks to the \textbf{NRS\_OLR} term,  \textbf{JOLR-DA} displays $44.48\%$ average accuracy and improves \textbf{JDA} by 7 points whereas the proposed full model, \textbf{DOLL-DA}, achieves the novel state of the art performance with $48.23\%$ average accuracy and improves \textbf{CDDA+} by $11.07$ points, \textbf{JOLR-DA} by roughly 4 points, and the former state of the art performance achieved by \textbf{DAH}  with a margin of $2.69$ points.

\subsubsection{\textbf{Experiments on the USPS+MNIST Data Set}}
\label{subsubsection: experiments on the UPS+MNIST Datasets}
\begin{figure*}[h!]
	\centering
	\includegraphics[width=0.8\linewidth]{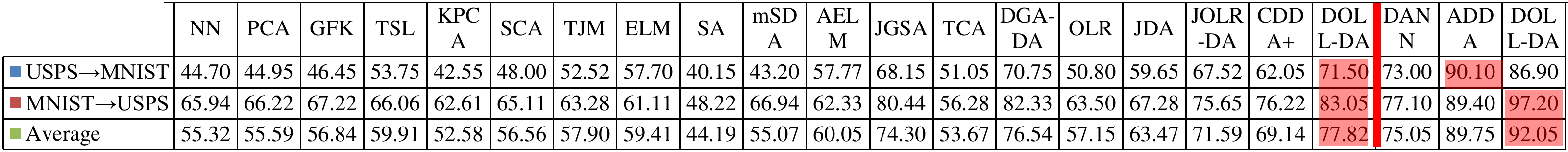}
	\caption { Accuracy${\rm{\% }}$ on the USPS+MNIST Images Dataset.} 
	\label{fig:accUSPS}
\end{figure*}

The USPS+MNIST dataset displays different writing styles between source and target. In Fig.\ref{fig:accUSPS}, the left columns of the red vertical bar report the experimental results using shallow features, whereas the right columns of the red bar display the results of the methods using deep features. Once more, the proposed \textbf{DOLL-DA} along with its partial models display the same behavior as in the previous experiences. Using shallow features, \textbf{CDDA+} and \textbf{JOLR-DA} demonstrate the effectiveness of the discriminative force terms and \textbf{NRS\_OLR} term. With $69.14\%$ and $71.59\%$ average accuracy, \textbf{CDDA+} and \textbf{JOLR-DA} improve the baseline \textbf{JDA} by 5 and 8 points, respectively.  When  jointly optimizing the discriminative force terms and \textbf{NRS\_OLR} term, the proposed final model, \textbf{DOLL-DA}, further boosts the performance  of the baseline \textbf{JDA} with a  margin as high as 14 points  and set a novel state of the art performance with $77.82\%$ average accuracy when shallow features are used. 

In right part of the red vertical bar in Fig.\ref{fig:accUSPS}, we compare our approach with deep network-based methods, \textit{i.e.}, \textbf{ADDA} and \textbf{DANN}, which search a common latent subspace for minimizing the source and target representation divergence through the popular adversarial learning, and improve the performance of traditional \textbf{DA} methods. As can be seen in Fig.\ref{fig:accUSPS}, the proposed \textbf{DOLL-DA} using shallow features already surpasses \textbf{DANN} by 2 points. When using deep features, \textit{i.e.}, those from LeNet \cite{lecun1998gradient} as in  \cite{rozantsev2018beyond}, \textbf{DOLL-DA} demonstrates its effectiveness once more and display a novel state of the art performance of $92.05\%$ average accuracy.



\subsubsection{\textbf{Large-Scale Experiments on the SVHN-MNIST Data Set}}
\label{subsubsection:Experiments on the SVHN-MNIST dataset}

\begin{figure}[h!]
	\centering
	\includegraphics[width=1\linewidth]{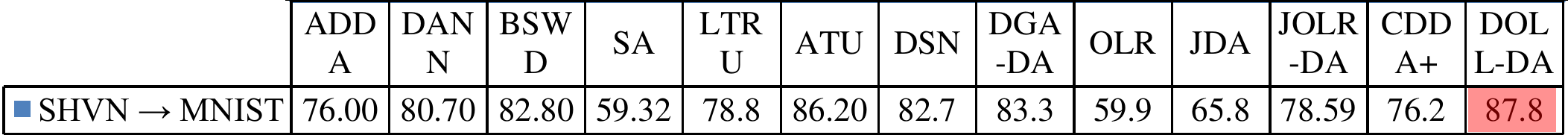}
	\caption { Accuracy${\rm{\% }}$ on the SVHN-MNIST Images Dataset.} 
	\label{fig:accSVHN}
\end{figure}

Differnt with the other datasets, SVHN-MNIST is compsed of 50k and 20k digit images from two very different domains, respectively, thereby generating a large scale \textbf{DA} benchmark.  Following the experiment setting of previous research\cite{luo2020discriminative},  Fig.\ref{fig:accSVHN} shows the experimental results. Our proposed \textbf{DOLL-DA} along with its partial models display the same patterns on this large scale benchmark as in the previous experiments. In ignoring the domain shifts,\textbf{OLR} achieves a poor performance.  Both \textbf{CDDA+} and \textbf{JOLR-DA} improve \textbf{JDA} with a large margin. In accounting jointly for the discriminative force and \textbf{NRS\_ORL}, \textbf{DOLL-DA} boosts the baseline \textbf{JDA} by $22$ points and surpasses \textbf{DGA-DA}, the former state of the art DA method on this dataset, by $4.5$ points. 



\subsubsection{\textbf{Experiments on the Office+Caltech-256 Data Sets}}
\label{subsubsection:Experiments on the Office+Caltech-256 Data Sets}

\begin{figure}[h!]
	\centering
	\includegraphics[width=1\linewidth]{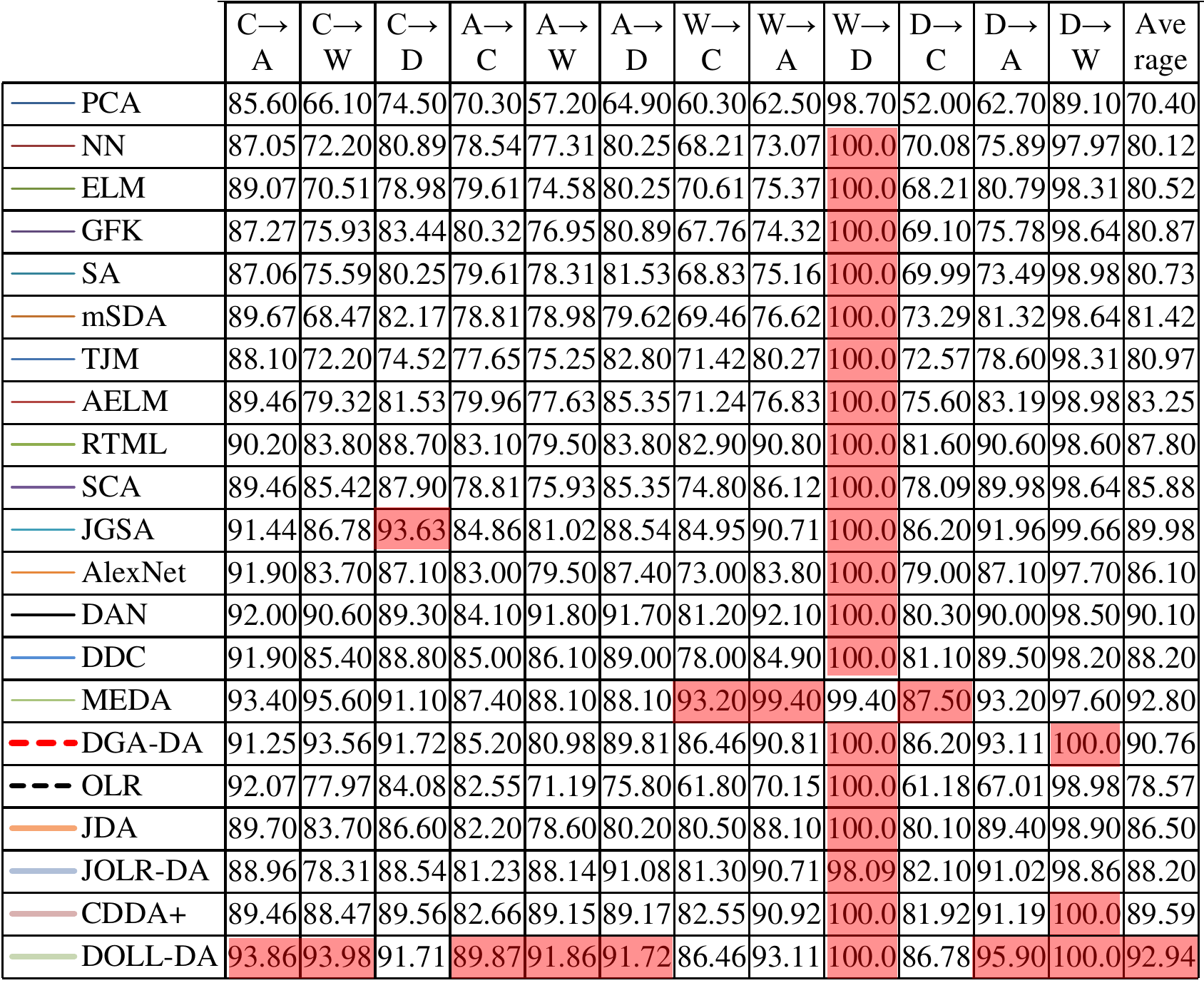}
	\caption { Accuracy${\rm{\% }}$ on the Office+Caltech Images with DeCAF6 Features.} 
	\label{fig:accDO}
\end{figure}

Fig.\ref{fig:accDO} and Fig.\ref{fig:accSO} synthesize the experimental results in comparison with the state of the art when deep features (\textit{i.e.}, DeCAF6 features) and classic shallow features (\textit{i.e.}, SURF features)   are used, respectively.

\begin{itemize}
	\item  As can be seen in Fig.\ref{fig:accSO}, using shallow features,  \textbf{CDDA+} and \textbf{JOLR-DA} improve the baseline \textbf{JDA} by 2 and 1 points, respectively, thanks to the discriminative force term and \textbf{NRS\_OLR} term introduced into the baseline model. Taking them together, our final model \textbf{DOLL-DA} further improves \textbf{JDA} by 4 points and achieves $51.18‰$ average accuracy which is in par with the previous state the art method, namely \textbf{JGSA}, with its $50.58\%$ average accuracy. It is worth noting that,  \textbf{JGSA} also suggests aligning data both statistically, discriminatively and geometrically, and corroborates data geometry aware DA approach, and achieves very good performance on this dataset.
	
		
	\begin{figure}[h!]
		\centering
		\includegraphics[width=1\linewidth]{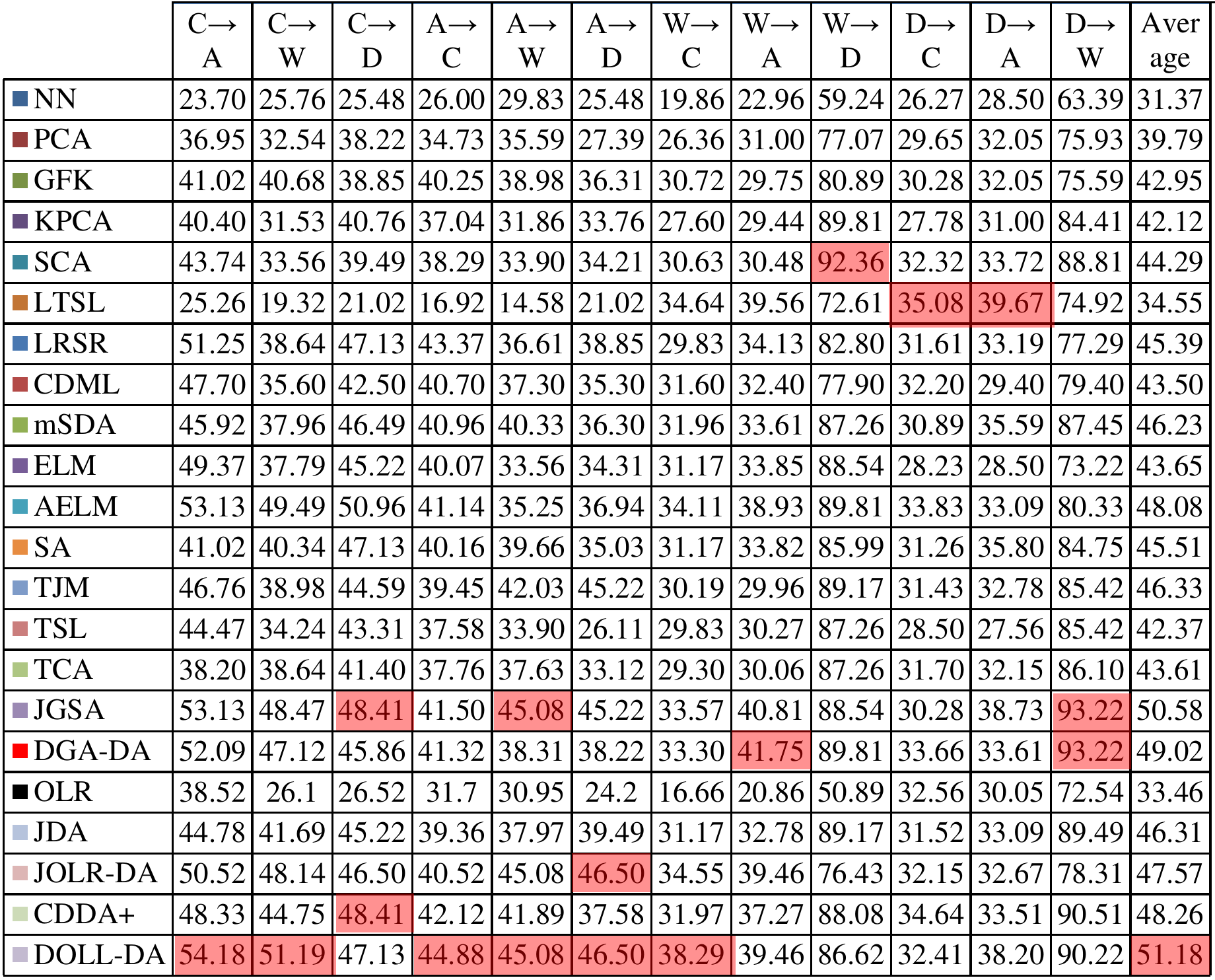}
		\caption { Accuracy${\rm{\% }}$ on the Office+Caltech Images with SURF-BoW Features.} 
		\label{fig:accSO}
	\end{figure}

	\item Fig.\ref{fig:accDO} compares the proposed DA method using deep features \textit{w.r.t.}   the state of the art, in particular end-to-end deep learning-based DA methods. As can be seen in Fig.\ref{fig:accDO}, the use of deep features has enabled impressive accuracy improvement over shallow features. Simple baseline methods, \textit{e.g.}, \textbf{NN}, \textbf{PCA}, see their accuracy soared by roughly 40 points, demonstrating the power of deep learning paradigm. Our proposed DA method also takes advantage of this jump and sees its accuracy soared from 48.26 to 89.59 for \textbf{CDDA+}, from 47.57 to 88.20 for \textbf{JOLR-DA},   and from 50.78 to 91.65 for \textbf{DOLL-DA}. As for shallow features, \textbf{CDDA+} and \textbf{JOLR-DA} improve \textbf{JDA} by roughly 3 and 2 points, respectively, while the final model, \textbf{DOLL-DA}, displays the best average accuracy of $92.94\%$ in par with $92.80\%$ displayed by \textbf{MEDA}.

\end{itemize}

\subsection{\textbf{Empirical Analysis}}
\label{Empirical Analysis}

Despite the proposed \textbf{DOLL-DA} displays state of the art performance over 49 DA tasks through 8 datasets except for the \textbf{COIL} dataset where it achieves the second best performance, an important question is how fast the proposed method converges (sect.\ref{Convergence analysis}) as well as its sensitivity \textit{w.r.t.} its hyper-parameters (Sect.\ref{Variants of Parameters}). Additionally, we are curious about how well  \textbf{DOLL-DA} performs in changing the base classifier (sect.\ref{Variants of base classifier:}) which is required for the formulation of the \textit{repulsive force} terms to enhance data discriminativeness, as well as the  \textbf{DOLL-DA} leveraging random initialization (sect.\ref{Random Initialization}) instead of using the base classifier for optimization. Furthermore, in analyzing the generalization capacity (sect.\ref{Generalization Capability}) of \textbf{DOLL-DA}, we evaluate the  performance of \textbf{DOLL-DA} with unseen target data for detail exploration.




\subsubsection{\textbf{Sensitivity of the proposed DOLL-DA \textit{w.r.t.} to hyper-parameters}}
\label{Variants of Parameters} Three hyper-parameters, namely  $k$, $\beta$ and $\alpha$, are introduced in the proposed methods.

\begin{figure}[h!]
	\begin{center}
		\begin{tabular}{c}
			\includegraphics[width=1\linewidth]{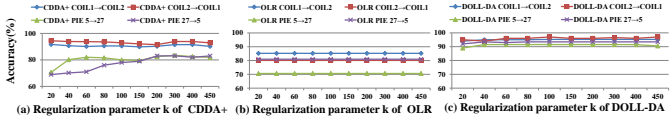}
		\end{tabular}
	\end{center}
	\vspace{-10pt} 
	\caption {Sensitivity analysis of the proposed methods:  (a) accuracy \textit{w.r.t.} subspace dimension  $k$ of \textbf{CDDA+}; (b)accuracy \textit{w.r.t.} subspace dimension $k$ of \textbf{JOLR-DA};
		(c) accuracy \textit{w.r.t.} subspace dimension $k$ of \textbf{DOLL-DA}.
		Three datasets are used, \textit{i.e.}, COIL1, COIL2 and PIE.} 
	\label{fig:k}
\end{figure}

\textit{Dimensionality analysis:}   $k$ is the dimension of the searched  shared latent feature subspace between the source and target domain. In Fig.\ref{fig:k}, we plot the classification accuracies of the proposed DA method \textit{w.r.t} different values of $k$ on the  \textbf{COIL} and \textbf{PIE} datasets. As shown in Fig.\ref{fig:k}, the subspace dimensionality $k$ varies with $k  \in \{20,40,60,80,100,150,200,300,400,450\}$, yet the proposed 3 DA variants, namely, \textbf{CDDA+}, \textbf{JOLR-DA} and \textbf{DOLL-DA}, remain stable \textit{w.r.t.}  a wide range of with $k \in \{ 40 \le k  \le 400\} $. It can be seen that both \textbf{JOLR-DA} and \textbf{DOLL-DA} display better robustness than \textbf{CDDA+}  \textit{w.r.t.} the variation of $k$, thereby suggesting the effectiveness of the \textbf{NRS\_OLR} term in the search of the global minimization. Obviously, the larger is $k$ the better the shared subspace can afford complex data distributions, but at the cost of increased computation complexity as highlighted in sect.\ref{subsection:Time Complexity Analysis} on time complexity analysis. In our experiments,  we set $k =300$ to balance the efficiency and accuracy.

\begin{figure}[h!]
	\centering
	\includegraphics[width=1\linewidth]{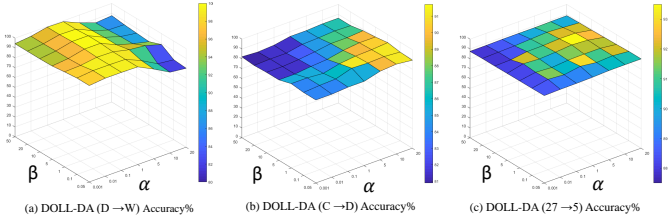}
	\caption { The classification accuracies of the proposed \textbf{DOLL-DA} method vs. the parameters $\alpha $ and $\beta $ on the selected three cross domains data sets, with $k$ held fixed at $300$.} 
	\label{fig:para}
\end{figure}

\textit{\textbf{Sensitivity of $\alpha$ and $\beta$:}} 
$\alpha $ and $\beta $ as defined in Eq.(\ref{eq:opti}) are the major hyper-parameters of the proposed \textbf{DOLL-DA}. While  $\alpha $ aims to regularize the projection matrix $A$ to avoid over-fitting the chosen shared feature subspace \textit{w.r.t.} both source and target domain data, $\beta$ as expressed in Eq.(\ref{eq:RLR}) controls the dimensionality of class dependent data manifold in the searched shared feature subspace, or in other words the sparsity level of the linear combination of the projected features to regress the class label. We study the sensitivity of the proposed \textbf{DOLL-DA} method with a wide range of parameter values, \textit{i.e.}, $\alpha  = (0.001,0.01,0.1,1,10,20,50)$ and $\beta  = (0.05,0.1,1,5,10,100,200)$. We plot in Fig.\ref{fig:para} the results on \emph{D} $\rightarrow$ \emph{W}, \emph{C} $\rightarrow$ \emph{D} $ and $ \emph{PIE-27} $\rightarrow$ \emph{PIE-5} datasets on the proposed \textbf{DOLL-DA} with $k$ held fixed at $300$. As can be seen from Fig.\ref{fig:para}, the proposed \textbf{DOLL-DA} displays its stability as the resultant classification  accuracies remain roughly the same despite  a wide range of $\alpha $ and $\beta $ values.

\subsubsection{\textbf{Convergence analysis}}
\label{Convergence analysis}

\begin{figure}[h!]
	\centering
	\includegraphics[width=1\linewidth]{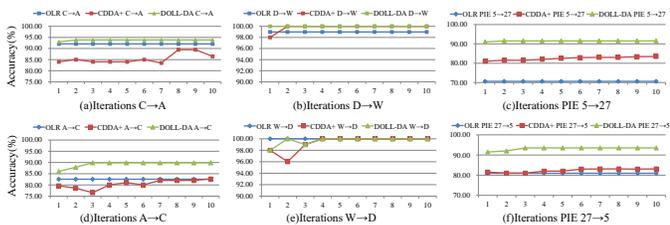}
	\caption {Convergence analysis using 6 cross-domain image classification tasks on Office+Caltech256 and PIE datasets. (accuracy w.r.t $\#$iterations) } 
	\label{fig:accITER}
\end{figure}

In Fig.\ref{fig:accITER}, we further perform  convergence analysis of the proposed \textbf{DOLL-DA} along with its partial models, \textit{i.e.},  \textbf{CDDA+} and \textbf{JOLR}, using the \textbf{DeCAF6} features on the \textbf{Office+Caltech} datasets and pixel value features on the \textbf{PIE} dataset. We aim to disclose how fast the proposed methods  achieve their best performance \textit{w.r.t.} the number of iterations $T$.   Fig.\ref{fig:accITER} reports 6 cross DA experiments ( \emph{C} $\rightarrow$ \emph{A}, \emph{D} $\rightarrow$ \emph{W} ...  \emph{PIE-27} $\rightarrow$ \emph{PIE-5}  ) with the number of iterations $T = (1,2,3,4,5,6,7,8,9,10)$. As shown in Fig.\ref{fig:accITER},  \textbf{CDDA+}, \textbf{JOLR-DA} and \textbf{DOLL-DA} converge within 3$ \sim $5 iterations during optimization, but \textbf{JOLR-DA} and \textbf{DOLL-DA} seem to converge even faster with a better accuracy, thanks to the \textbf{NRS\_OLR} term introduced in our DA model.


\subsubsection{\textbf{Impact of the base classifier}}
\label{Variants of base classifier:}

The \textit{repulsive force} term across domain as formulated in sect.\ref{Repulsing interclass data for discriminative DA} requires using pseudo labels for the target domain data. Therefore, the quality of these pseudo labels could have much impact on the effectiveness of data discriminativeness. In sect.\ref{Solving the model}, our model is initialized  using  pseudo labels in solving Eq.(\ref{eq:S2}),  which boils down to \textbf{JDA}. The inference of the pseudo labels on the target domain data requires a base classifier trained using the labeled source domain data. We test the sensitivity of the proposed \textbf{DOLL-DA} \textit{w.r.t.} the base classifier using two popular classifiers, \textit{i.e.}, NN and SVM.  Fig.\ref{fig:rand} shows that \textbf{DOLL-DA-NN}(92.94\%) and \textbf{DOLL-DA-SVM}(91.51\%) achieve almost the same performance. This result suggests that our \textit{repulsive force} term in the proposed \textbf{DOLL-DA} displays a certain level of robustness \textit{w.r.t.} the choice of the base classifier.

\subsubsection{\textbf{Random Label Initialization}}
\label{Random Initialization}

Going one step further \textit{w.r.t.} to the experiment in sect.\ref{Variants of base classifier:}, an interesting question is how \textbf{DOLL-DA} behaves with randomly initialized pseudo labels at its first iteration as well as its convergence efficiency in this specific experiment setting.  

\begin{figure}[h!]
	\centering
	\includegraphics[width=1\linewidth]{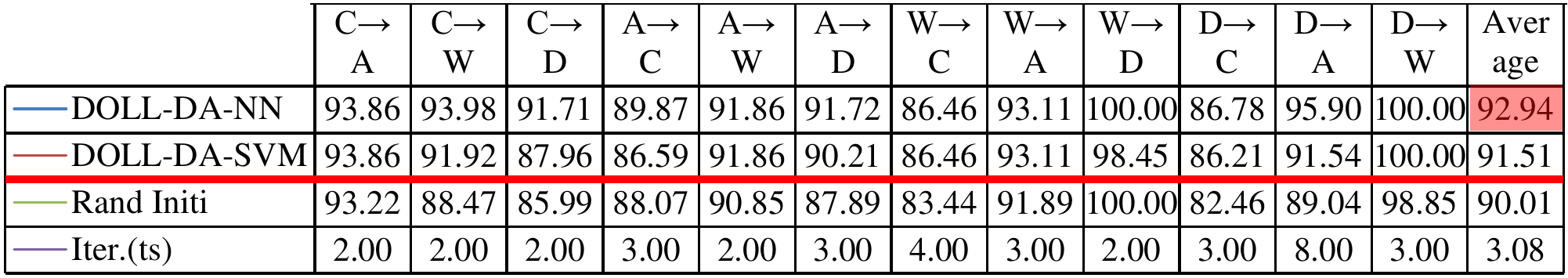}
	\caption {Convergence analysis using 12 cross-domain image classification tasks on Office+Caltech256 datasets with DeCAF6 Features. (accuracy w.r.t $\#$iterations) } 
	\label{fig:rand}
\end{figure}

In this setting, \textbf{DOLL-DA} makes use of \textbf{randomly initialized} labels for the target domain data instead of solving Eq.(\ref{eq:S2}) at its first iteration. As shown in Fig.\ref{fig:rand}, \textbf{DOLL-DA} still achieves 90.01\% accuracy on the \textbf{Office-Caltech256} dataset, thus only slightly below \textbf{DOLL-DA-NN}, and converges on average at 3.08 average iterations. This result further supports the robustness of the designed \textit{repulsive force} term regularized by the \textbf{NRS\_OLR} term in the search of the optimized shared features subspace.

\subsubsection{\textbf{Stability \textit{w.r.t.} target domain data}}
\label{Generalization Capability}

\begin{figure}[h!]
	\centering
	\includegraphics[width=0.9\linewidth]{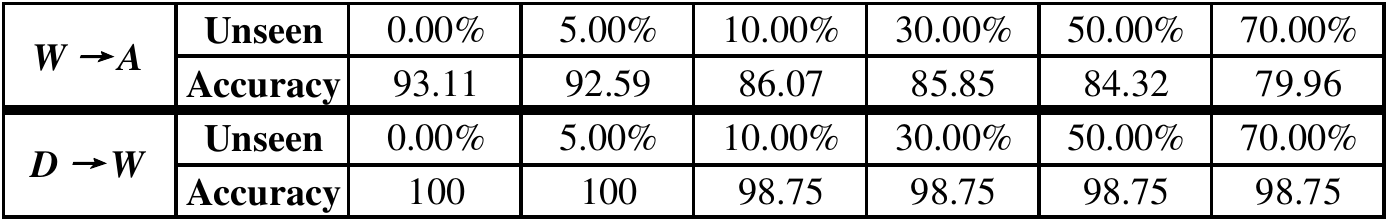}
	\caption {Unseen data sensitivity using 2 cross-domain image classification tasks on Office+Caltech256 datasets with DeCAF6 Features. } 
	\label{fig:unseen}
\end{figure}

We benchamrk the stability of the proposed \textbf{DOLL-DA} \textit{w.r.t.} the quantity of target domain data used for training. Specifically, we carry out two additional experiments for 2 DA tasks using the Office-Caltech256 dataset, \textit{i.e.},   \emph{W} $\rightarrow$ \emph{A}, \emph{D} $\rightarrow$ \emph{W} ,  keeping 5${\rm{\% }}$ (10${\rm{\% }}$,30${\rm{\% }}$,50${\rm{\% }}$,70${\rm{\% }}$, resp.) of target domain data from being used as auxiliary data in training. Results are reported in Fig.\ref{fig:unseen}. As can be seen there, when all target domain data are used in training the proposed \textbf{DOLL-DA}, \textit{i.e.}, column $0.00\%$, \textbf{DOLL-DA} displays $93.11\%$ and $100\%$ accuracy for \emph{W} $\rightarrow$ \emph{A} and \emph{D} $\rightarrow$ \emph{W} DA tasks, respectively. Now when more and more target domain data are kept from being used in training, passing from 5${\rm{\% }}$ through 70${\rm{\% }}$, \textbf{DOLL-DA} proves quite stable on the \emph{D} $\rightarrow$ \emph{W} task but decreases constantly to reach $79.96\%$ on the \emph{W} $\rightarrow$ \emph{A} task. However, this result still proves the usefulness of the proposed DA method when only $30\%$ target domain data are used as auxiliary data in training, given the fact that the baseline NN only displays $73.07\%$ accuracy as shown in Fig.\ref{fig:accDO}. The performance difference between these two DA tasks can be explained by their inherent difficulties. The DA task \emph{D} $\rightarrow$ \emph{W} is to generalize a classifier trained from the source domain, \textit{i.e.}, labeled images in the DSLR domain, thus with much background, to the target domain, \textit{i.e.}, images in the Webcam domain, much simpler because devoid of the background, and the simple baseline NN already achieves $97.97\%$ accuracy as shown in Fig.\ref{fig:accDO}. On the other side, the DA task \emph{W} $\rightarrow$ \emph{A} does exactly the contrary and needs to generalize a classifier trained from the source domain, \textit{i.e.}, labeled images in the Webcam domain, thus without background, to a much more complicated target domain, \textit{i.e.}, images in the Amazone domain with arbitrary background, and the simple baseline NN only achieves $73.07\%$ accuracy as shown in Fig.\ref{fig:accDO}.   


\section{Conclusion}
\label{Conclusion}
We have proposed in this paper a novel unsupervised DA method, namely \textbf{D}iscriminative Noise Robust Sparse \textbf{O}rthogonal \textbf{L}abe\textbf{l} Regression-based \textbf{D}omain \textbf{A}daptation (\textbf{DOLL-DA}), which simultaneously optimizes the three terms of the upper error bound of a learned classifier on the target domain in aligning discriminatively data distributions through a \textit{repulse force} term while orthogonally regressing data labels within the shared feature subspace. Furthermore, the proposed model explicitly accounts for data outliers to avoid negative transfer and introduces the property of sparsity when regressing data labels. Comprehensive experiments using the standard benchmark in DA show the effectiveness of the proposed method which consistently outperform state of the art DA methods. Future work includes embedding of the proposed \textbf{DOLL-DA} into the paradigm of deep learning and considers the setting of online learning where target domain data only arrives sequentially one after another.

\ifCLASSOPTIONcaptionsoff
  \newpage
\fi

\bibliographystyle{plain}
\bibliography{egbib}

\begin{thebibliography}{10}

\bibitem{ben2010theory}
Shai Ben-David, John Blitzer, Koby Crammer, Alex Kulesza, Fernando Pereira, and
  Jennifer~Wortman Vaughan.
\newblock A theory of learning from different domains.
\newblock {\em Machine learning}, 79(1):151--175, 2010.

\bibitem{ben2007analysis}
Shai Ben-David, John Blitzer, Koby Crammer, and Fernando Pereira.
\newblock Analysis of representations for domain adaptation.
\newblock In {\em Advances in neural information processing systems}, pages
  137--144, 2007.

\bibitem{borgwardt2006integrating}
Karsten~M Borgwardt, Arthur Gretton, Malte~J Rasch, Hans-Peter Kriegel,
  Bernhard Sch{\"o}lkopf, and Alex~J Smola.
\newblock Integrating structured biological data by kernel maximum mean
  discrepancy.
\newblock {\em Bioinformatics}, 22(14):e49--e57, 2006.

\bibitem{bousmalis2016domain}
Konstantinos Bousmalis, George Trigeorgis, Nathan Silberman, Dilip Krishnan,
  and Dumitru Erhan.
\newblock Domain separation networks.
\newblock In {\em Advances in Neural Information Processing Systems}, pages
  343--351, 2016.

\bibitem{DBLP:journals/corr/abs-1206-4683}
Minmin Chen, Zhixiang~Eddie Xu, Kilian~Q. Weinberger, and Fei Sha.
\newblock Marginalized denoising autoencoders for domain adaptation.
\newblock {\em CoRR}, abs/1206.4683, 2012.

\bibitem{courty2017joint}
Nicolas Courty, R{\'e}mi Flamary, Amaury Habrard, and Alain Rakotomamonjy.
\newblock Joint distribution optimal transportation for domain adaptation.
\newblock In {\em Advances in Neural Information Processing Systems}, pages
  3733--3742, 2017.

\bibitem{courty2017optimal}
Nicolas Courty, R{\'e}mi Flamary, Devis Tuia, and Alain Rakotomamonjy.
\newblock Optimal transport for domain adaptation.
\newblock {\em IEEE transactions on pattern analysis and machine intelligence},
  39(9):1853--1865, 2017.

\bibitem{DBLP:journals/tip/DingF17}
Zhengming Ding and Yun Fu.
\newblock Robust transfer metric learning for image classification.
\newblock {\em {IEEE} Trans. Image Processing}, 26(2):660--670, 2017.

\bibitem{DBLP:journals/tnn/DingF18}
Zhengming Ding and Yun Fu.
\newblock Robust multiview data analysis through collective low-rank subspace.
\newblock {\em {IEEE} Trans. Neural Netw. Learning Syst.}, 29(5):1986--1997,
  2018.

\bibitem{donahue2013semi}
Jeff Donahue, Judy Hoffman, Erik Rodner, Kate Saenko, and Trevor Darrell.
\newblock Semi-supervised domain adaptation with instance constraints.
\newblock In {\em Proceedings of the IEEE conference on computer vision and
  pattern recognition}, pages 668--675, 2013.

\bibitem{DBLP:conf/icml/DonahueJVHZTD14}
Jeff Donahue, Yangqing Jia, Oriol Vinyals, Judy Hoffman, Ning Zhang, Eric
  Tzeng, and Trevor Darrell.
\newblock Decaf: {A} deep convolutional activation feature for generic visual
  recognition.
\newblock In {\em Proceedings of the 31th International Conference on Machine
  Learning, {ICML} 2014, Beijing, China, 21-26 June 2014}, pages 647--655,
  2014.

\bibitem{DBLP:conf/iccv/FernandoHST13}
Basura Fernando, Amaury Habrard, Marc Sebban, and Tinne Tuytelaars.
\newblock Unsupervised visual domain adaptation using subspace alignment.
\newblock In {\em {IEEE} International Conference on Computer Vision, {ICCV}
  2013, Sydney, Australia, December 1-8, 2013}, pages 2960--2967, 2013.

\bibitem{fiori2005formulation}
Simone Fiori.
\newblock Formulation and integration of learning differential equations on the
  stiefel manifold.
\newblock {\em IEEE transactions on neural networks}, 16(6):1697--1701, 2005.

\bibitem{ganin2016domain}
Yaroslav Ganin, Evgeniya Ustinova, Hana Ajakan, Pascal Germain, Hugo
  Larochelle, Fran{\c{c}}ois Laviolette, Mario Marchand, and Victor Lempitsky.
\newblock Domain-adversarial training of neural networks.
\newblock {\em The Journal of Machine Learning Research}, 17(1):2096--2030,
  2016.

\bibitem{DBLP:journals/pami/GhifaryBKZ17}
Muhammad Ghifary, David Balduzzi, W.~Bastiaan Kleijn, and Mengjie Zhang.
\newblock Scatter component analysis: {A} unified framework for domain
  adaptation and domain generalization.
\newblock {\em {IEEE} Trans. Pattern Anal. Mach. Intell.}, 39(7):1414--1430,
  2017.

\bibitem{gong2012geodesic}
Boqing Gong, Yuan Shi, Fei Sha, and Kristen Grauman.
\newblock Geodesic flow kernel for unsupervised domain adaptation.
\newblock In {\em Computer Vision and Pattern Recognition (CVPR), 2012 IEEE
  Conference on}, pages 2066--2073. IEEE, 2012.

\bibitem{goodfellow2014generative}
Ian Goodfellow, Jean Pouget-Abadie, Mehdi Mirza, Bing Xu, David Warde-Farley,
  Sherjil Ozair, Aaron Courville, and Yoshua Bengio.
\newblock Generative adversarial nets.
\newblock In {\em Advances in neural information processing systems}, pages
  2672--2680, 2014.

\bibitem{gretton2007kernel}
Arthur Gretton, Karsten~M Borgwardt, Malte Rasch, Bernhard Sch{\"o}lkopf, and
  Alex~J Smola.
\newblock A kernel method for the two-sample-problem.
\newblock In {\em Advances in neural information processing systems}, pages
  513--520, 2007.

\bibitem{harandi2017dimensionality}
Mehrtash Harandi, Mathieu Salzmann, and Richard Hartley.
\newblock Dimensionality reduction on spd manifolds: The emergence of
  geometry-aware methods.
\newblock {\em IEEE transactions on pattern analysis and machine intelligence},
  2017.

\bibitem{herath2017learning}
Samitha Herath, Mehrtash Harandi, and Fatih Porikli.
\newblock Learning an invariant hilbert space for domain adaptation.
\newblock In {\em Proceedings of the IEEE Conference on Computer Vision and
  Pattern Recognition}, pages 3845--3854, 2017.

\bibitem{DBLP:conf/cvpr/HerathHP17}
Samitha Herath, Mehrtash~Tafazzoli Harandi, and Fatih Porikli.
\newblock Learning an invariant hilbert space for domain adaptation.
\newblock In {\em 2017 {IEEE} Conference on Computer Vision and Pattern
  Recognition, {CVPR} 2017, Honolulu, HI, USA, July 21-26, 2017}, pages
  3956--3965, 2017.

\bibitem{pmlr-v80-hoffman18a}
Judy Hoffman, Eric Tzeng, Taesung Park, Jun-Yan Zhu, Phillip Isola, Kate
  Saenko, Alexei Efros, and Trevor Darrell.
\newblock {C}y{CADA}: Cycle-consistent adversarial domain adaptation.
\newblock In Jennifer Dy and Andreas Krause, editors, {\em Proceedings of the
  35th International Conference on Machine Learning}, volume~80 of {\em
  Proceedings of Machine Learning Research}, pages 1989--1998,
  Stockholmsmässan, Stockholm Sweden, 10--15 Jul 2018. PMLR.

\bibitem{DBLP:journals/pami/Hull94}
Jonathan~J. Hull.
\newblock A database for handwritten text recognition research.
\newblock {\em {IEEE} Trans. Pattern Anal. Mach. Intell.}, 16(5):550--554,
  1994.

\bibitem{jhuo2012robust}
I-Hong Jhuo, Dong Liu, DT~Lee, and Shih-Fu Chang.
\newblock Robust visual domain adaptation with low-rank reconstruction.
\newblock In {\em Computer Vision and Pattern Recognition (CVPR), 2012 IEEE
  Conference on}, pages 2168--2175. IEEE, 2012.

\bibitem{karbalayghareh2018optimal}
Alireza Karbalayghareh, Xiaoning Qian, and Edward~R Dougherty.
\newblock Optimal bayesian transfer learning.
\newblock {\em IEEE Transactions on Signal Processing}, 66(14):3724--3739,
  2018.

\bibitem{kifer2004detecting}
Daniel Kifer, Shai Ben-David, and Johannes Gehrke.
\newblock Detecting change in data streams.
\newblock In {\em Proceedings of the Thirtieth international conference on Very
  large data bases-Volume 30}, pages 180--191. VLDB Endowment, 2004.

\bibitem{krizhevsky2012imagenet}
Alex Krizhevsky, Ilya Sutskever, and Geoffrey~E Hinton.
\newblock Imagenet classification with deep convolutional neural networks.
\newblock In {\em Advances in neural information processing systems}, pages
  1097--1105, 2012.

\bibitem{lecun1998gradient}
Yann LeCun, L{\'e}on Bottou, Yoshua Bengio, and Patrick Haffner.
\newblock Gradient-based learning applied to document recognition.
\newblock {\em Proceedings of the IEEE}, 86(11):2278--2324, 1998.

\bibitem{li2019locality}
Jingjing Li, Mengmeng Jing, Ke~Lu, Lei Zhu, and Heng~Tao Shen.
\newblock Locality preserving joint transfer for domain adaptation.
\newblock {\em IEEE Transactions on Image Processing}, 28(12):6103--6115, 2019.

\bibitem{liang2018aggregating}
Jian Liang, Ran He, Zhenan Sun, and Tieniu Tan.
\newblock Aggregating randomized clustering-promoting invariant projections for
  domain adaptation.
\newblock {\em IEEE Transactions on Pattern Analysis and Machine Intelligence},
  2018.

\bibitem{liu2019homologous}
Youfa Liu, Weiping Tu, Bo~Du, Lefei Zhang, and Dacheng Tao.
\newblock Homologous component analysis for domain adaptation.
\newblock {\em IEEE Transactions on Image Processing}, 29:1074--1089, 2019.

\bibitem{long2015learning}
Mingsheng Long, Yue Cao, Jianmin Wang, and Michael~I Jordan.
\newblock Learning transferable features with deep adaptation networks.
\newblock In {\em ICML}, pages 97--105, 2015.

\bibitem{long2013transfer}
Mingsheng Long, Jianmin Wang, Guiguang Ding, Jiaguang Sun, and Philip~S Yu.
\newblock Transfer feature learning with joint distribution adaptation.
\newblock In {\em Proceedings of the IEEE International Conference on Computer
  Vision}, pages 2200--2207, 2013.

\bibitem{DBLP:conf/cvpr/LongWDSY14}
Mingsheng Long, Jianmin Wang, Guiguang Ding, Jiaguang Sun, and Philip~S. Yu.
\newblock Transfer joint matching for unsupervised domain adaptation.
\newblock In {\em 2014 {IEEE} Conference on Computer Vision and Pattern
  Recognition, {CVPR} 2014, Columbus, OH, USA, June 23-28, 2014}, pages
  1410--1417, 2014.

\bibitem{DBLP:conf/icml/LongZ0J17}
Mingsheng Long, Han Zhu, Jianmin Wang, and Michael~I. Jordan.
\newblock Deep transfer learning with joint adaptation networks.
\newblock In {\em Proceedings of the 34th International Conference on Machine
  Learning, {ICML} 2017, Sydney, NSW, Australia, 6-11 August 2017}, pages
  2208--2217, 2017.

\bibitem{lu2018embarrassingly}
Hao Lu, Chunhua Shen, Zhiguo Cao, Yang Xiao, and Anton van~den Hengel.
\newblock An embarrassingly simple approach to visual domain adaptation.
\newblock {\em IEEE Transactions on Image Processing}, 2018.

\bibitem{Ying_TL_Survey_10.1145/3379344}
Ying Lu, Lingkun Luo, Di~Huang, Yunhong Wang, and Liming Chen.
\newblock Knowledge transfer in vision recognition: A survey.
\newblock {\em ACM Comput. Surv.}, 53(2), April 2020.

\bibitem{DBLP:journals/corr/abs-1712-10042}
Lingkun Luo, Liming Chen, Shiqiang Hu, Ying Lu, and Xiaofang Wang.
\newblock Discriminative and geometry aware unsupervised domain adaptation.
\newblock {\em CoRR}, abs/1712.10042, 2017.

\bibitem{luo2020discriminative}
Lingkun Luo, Liming Chen, Shiqiang Hu, Ying Lu, and Xiaofang Wang.
\newblock Discriminative and geometry-aware unsupervised domain adaptation.
\newblock {\em IEEE Transactions on Cybernetics}, 2020.

\bibitem{DBLP:journals/corr/LuoWHC17}
Lingkun Luo, Xiaofang Wang, Shiqiang Hu, and Liming Chen.
\newblock Robust data geometric structure aligned close yet discriminative
  domain adaptation.
\newblock {\em CoRR}, abs/1705.08620, 2017.

\bibitem{nie2014optimal}
Feiping Nie, Jianjun Yuan, and Heng Huang.
\newblock Optimal mean robust principal component analysis.
\newblock In {\em International conference on machine learning}, pages
  1062--1070, 2014.

\bibitem{pan2011domain}
Sinno~Jialin Pan, Ivor~W Tsang, James~T Kwok, and Qiang Yang.
\newblock Domain adaptation via transfer component analysis.
\newblock {\em IEEE Transactions on Neural Networks}, 22(2):199--210, 2011.

\bibitem{pan2010survey}
Sinno~Jialin Pan and Qiang Yang.
\newblock A survey on transfer learning.
\newblock {\em IEEE Transactions on knowledge and data engineering},
  22(10):1345--1359, 2010.

\bibitem{Busto_2017_ICCV}
Pau Panareda~Busto and Juergen Gall.
\newblock Open set domain adaptation.
\newblock In {\em The IEEE International Conference on Computer Vision (ICCV)},
  Oct 2017.

\bibitem{7078994}
V.~M. Patel, R.~Gopalan, R.~Li, and R.~Chellappa.
\newblock Visual domain adaptation: A survey of recent advances.
\newblock {\em IEEE Signal Processing Magazine}, 32(3):53--69, May 2015.

\bibitem{pei2018multi}
Zhongyi Pei, Zhangjie Cao, Mingsheng Long, and Jianmin Wang.
\newblock Multi-adversarial domain adaptation.
\newblock In {\em Thirty-Second AAAI Conference on Artificial Intelligence},
  2018.

\bibitem{rozantsev2018beyond}
Artem Rozantsev, Mathieu Salzmann, and Pascal Fua.
\newblock Beyond sharing weights for deep domain adaptation.
\newblock {\em IEEE Transactions on Pattern Analysis and Machine Intelligence},
  2018.

\bibitem{DBLP:conf/icml/SaitoUH17}
Kuniaki Saito, Yoshitaka Ushiku, and Tatsuya Harada.
\newblock Asymmetric tri-training for unsupervised domain adaptation.
\newblock In {\em Proceedings of the 34th International Conference on Machine
  Learning, {ICML} 2017, Sydney, NSW, Australia, 6-11 August 2017}, pages
  2988--2997, 2017.

\bibitem{DBLP:journals/neco/ScholkopfSM98}
Bernhard Sch{\"{o}}lkopf, Alexander~J. Smola, and Klaus{-}Robert M{\"{u}}ller.
\newblock Nonlinear component analysis as a kernel eigenvalue problem.
\newblock {\em Neural Computation}, 10(5):1299--1319, 1998.

\bibitem{sener2016learning}
Ozan Sener, Hyun~Oh Song, Ashutosh Saxena, and Silvio Savarese.
\newblock Learning transferrable representations for unsupervised domain
  adaptation.
\newblock In {\em Advances in Neural Information Processing Systems}, pages
  2110--2118, 2016.

\bibitem{DBLP:journals/ijcv/ShaoKF14}
Ming Shao, Dmitry Kit, and Yun Fu.
\newblock Generalized transfer subspace learning through low-rank constraint.
\newblock {\em International Journal of Computer Vision}, 109(1-2):74--93,
  2014.

\bibitem{4967588}
S.~Si, D.~Tao, and B.~Geng.
\newblock Bregman divergence-based regularization for transfer subspace
  learning.
\newblock {\em IEEE Transactions on Knowledge and Data Engineering},
  22(7):929--942, July 2010.

\bibitem{si2010bregman}
Si~Si, Dacheng Tao, and Bo~Geng.
\newblock Bregman divergence-based regularization for transfer subspace
  learning.
\newblock {\em IEEE Transactions on Knowledge and Data Engineering},
  22(7):929--942, 2010.

\bibitem{sun2016return}
Baochen Sun, Jiashi Feng, and Kate Saenko.
\newblock Return of frustratingly easy domain adaptation.
\newblock In {\em AAAI}, volume~6, page~8, 2016.

\bibitem{sun2016deep}
Baochen Sun and Kate Saenko.
\newblock Deep coral: Correlation alignment for deep domain adaptation.
\newblock In {\em European Conference on Computer Vision}, pages 443--450.
  Springer, 2016.

\bibitem{tang2017visual}
Yuxing Tang, Josiah Wang, Xiaofang Wang, Boyang Gao, Emmanuel Dellandrea,
  Robert Gaizauskas, and Liming Chen.
\newblock Visual and semantic knowledge transfer for large scale
  semi-supervised object detection.
\newblock {\em IEEE Transactions on Pattern Analysis and Machine Intelligence},
  2017.

\bibitem{tzeng2017adversarial}
Eric Tzeng, Judy Hoffman, Kate Saenko, and Trevor Darrell.
\newblock Adversarial discriminative domain adaptation.
\newblock In {\em Computer Vision and Pattern Recognition (CVPR)}, volume~1,
  page~4, 2017.

\bibitem{DBLP:journals/corr/TzengHZSD14}
Eric Tzeng, Judy Hoffman, Ning Zhang, Kate Saenko, and Trevor Darrell.
\newblock Deep domain confusion: Maximizing for domain invariance.
\newblock {\em CoRR}, abs/1412.3474, 2014.

\bibitem{DBLP:journals/tcyb/UzairM17}
Muhammad Uzair and Ajmal~S. Mian.
\newblock Blind domain adaptation with augmented extreme learning machine
  features.
\newblock {\em {IEEE} Trans. Cybernetics}, 47(3):651--660, 2017.

\bibitem{venkateswara2017deep}
Hemanth Venkateswara, Jose Eusebio, Shayok Chakraborty, and Sethuraman
  Panchanathan.
\newblock Deep hashing network for unsupervised domain adaptation.
\newblock {\em arXiv preprint arXiv:1706.07522}, 2017.

\bibitem{DBLP:conf/aaai/WangWZX14}
Hao Wang, Wei Wang, Chen Zhang, and Fanjiang Xu.
\newblock Cross-domain metric learning based on information theory.
\newblock In {\em Proceedings of the Twenty-Eighth {AAAI} Conference on
  Artificial Intelligence, July 27 -31, 2014, Qu{\'{e}}bec City, Qu{\'{e}}bec,
  Canada.}, pages 2099--2105, 2014.

\bibitem{wang2018visual}
Jindong Wang, Wenjie Feng, Yiqiang Chen, Han Yu, Meiyu Huang, and Philip~S Yu.
\newblock Visual domain adaptation with manifold embedded distribution
  alignment.
\newblock In {\em 2018 ACM Multimedia Conference on Multimedia Conference},
  pages 402--410. ACM, 2018.

\bibitem{DBLP:journals/tip/XuFWLZ16}
Yong Xu, Xiaozhao Fang, Jian Wu, Xuelong Li, and David Zhang.
\newblock Discriminative transfer subspace learning via low-rank and sparse
  representation.
\newblock {\em {IEEE} Trans. Image Processing}, 25(2):850--863, 2016.

\bibitem{Zhang_2017_CVPR}
Jing Zhang, Wanqing Li, and Philip Ogunbona.
\newblock Joint geometrical and statistical alignment for visual domain
  adaptation.
\newblock In {\em The IEEE Conference on Computer Vision and Pattern
  Recognition (CVPR)}, July 2017.

\bibitem{zhao2019multi}
Sicheng Zhao, Bo~Li, Xiangyu Yue, Yang Gu, Pengfei Xu, Runbo Hu, Hua Chai, and
  Kurt Keutzer.
\newblock Multi-source domain adaptation for semantic segmentation.
\newblock In {\em Advances in Neural Information Processing Systems}, pages
  7285--7298, 2019.

\end{thebibliography}

	\vspace{-10pt} 
\begin{IEEEbiography}   [{\includegraphics[width=1in,height=1.25in,clip,keepaspectratio]{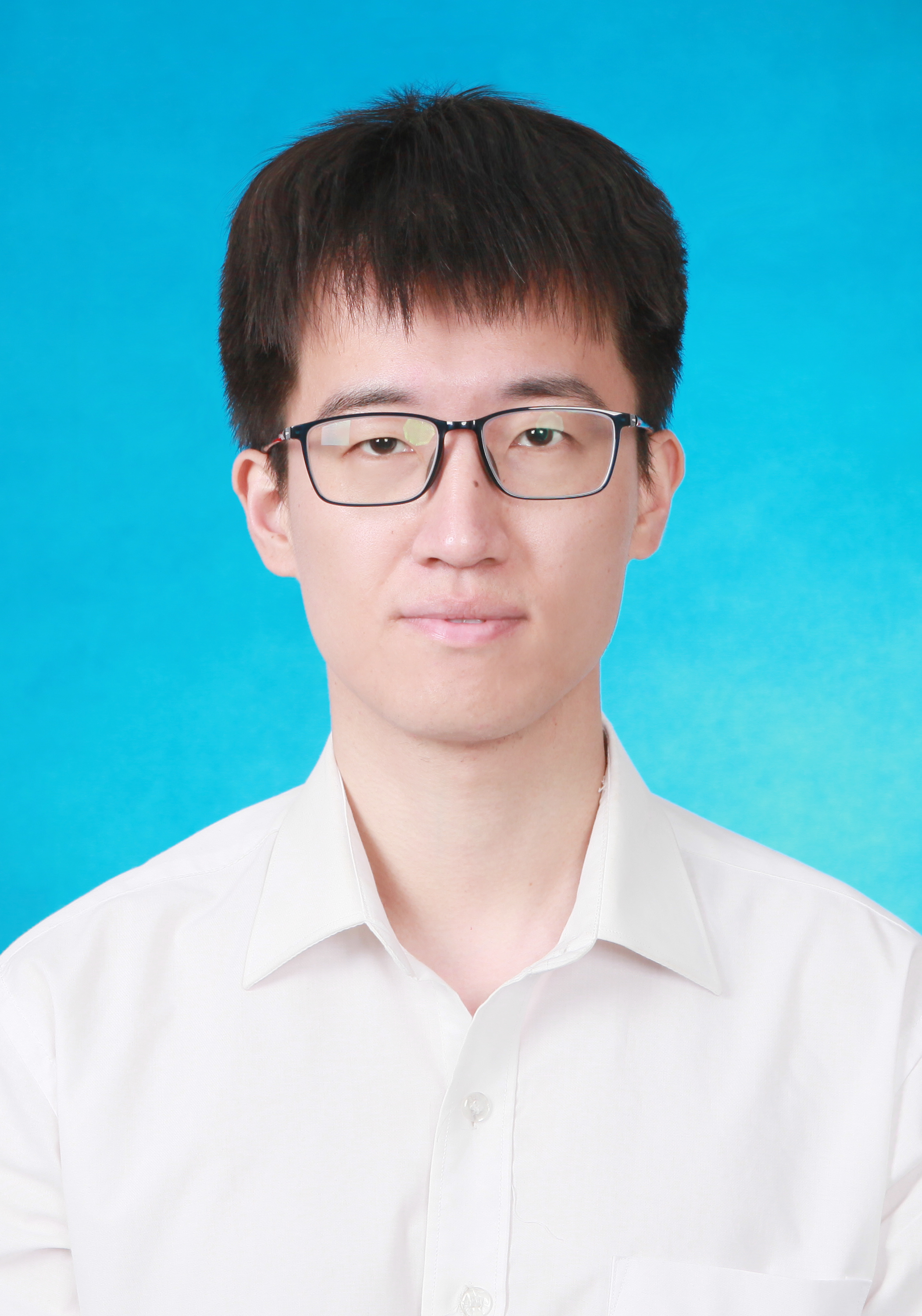}}]{Lingkun Luo}
	received his Ph.D degree in Shanghai Jiao Tong University. He had served as research assistant and postdoc in Ecole Centrale de Lyon, Department of Mathematics and Computer Science, and a member of LIRIS laboratory. Now, he is a postdoc in Shanghai Jiao Tong University. He has authored more
    than 20 research articles. His research interests include machine learning, pattern recognition and computer vision.
\end{IEEEbiography}
	\vspace{-5pt} 
\begin{IEEEbiography}[{\includegraphics[width=1in,height=1.25in,clip,keepaspectratio]{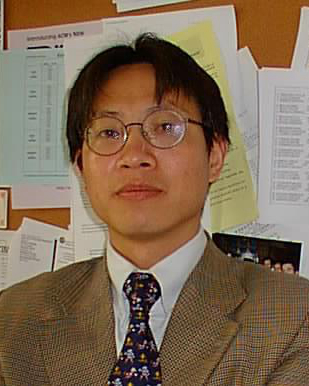}}]{Liming Chen} received the joint B.Sc. degree in mathematics and computer science from the University of Nantes, Nantes, France in 1984, and the M.Sc. and Ph.D. degrees in computer science from the University of Paris 6, Paris, France, in 1986 and 1989, respectively.

He first served as an Associate Professor with the Universit\'{e} de Technologie de Compi\`{e}gne, before joining \'Ecole Centrale de Lyon, \'Ecully, France, as a Professor in 1998,  where he leads an advanced research team on Computer Vision, Machine Learning and Multimedia. From 2001 to 2003, he also served as Chief Scientific Officer in a Paris-based company, Avivias, specializing in media asset management. In 2005, he served as Scientific Multimedia Expert for France Telecom R\&D China, Beijing, China. He was the Head of the Department of Mathematics and Computer Science, \'Ecole Centrale de Lyon from 2007 through 2016. His current research interests include computer vision, machine learning, image and and multimedia with a particular focus on robot vision and learning since 2016. Liming has over 300 publications and successfully supervised over 40 PhD students. He has been a grant holder for a number of research grants from EU FP program, French research funding bodies and local government departments. Liming has so far guest-edited 5 journal special issues. He is an associate editor for Eurasip Journal on Image and Video Processing and a senior IEEE member.
\end{IEEEbiography}

\begin{IEEEbiography}   [{\includegraphics[width=1in,height=1.25in,clip,keepaspectratio]{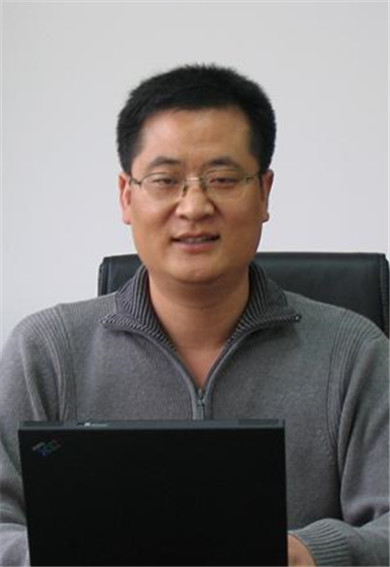}}]{Shiqiang Hu}
	received his PhD degree at Beijing Institute of Technology. He has over 150 publications and successfully supervised over 15 PhD students. Now, he is a full professor in Shanghai Jiao Tong University. His research interests include data fusion technology, image understanding, and nonlinear filter.
\end{IEEEbiography}

\end{CJK*}
\end{document}